\documentclass{article} 

\usepackage{iclr2025_conference,times}

\usepackage{wrapfig}
\usepackage{lipsum} 
\usepackage{wrapfig}

\usepackage{booktabs}      
\usepackage{multirow}      
\usepackage{amssymb}       
\usepackage{graphicx, subfigure}     

\usepackage{colortbl}
\usepackage{xcolor}

\usepackage{hyperref}

\usepackage{url}
\usepackage{multicol}
\usepackage{bm}
\usepackage{aliascnt}
\usepackage{adjustbox}
\usepackage{siunitx} 
\usepackage{booktabs}
\usepackage{multirow} 
\usepackage{enumitem}
\usepackage{booktabs} 
\usepackage{amssymb}  
\usepackage{amsmath}  
\usepackage{graphicx} 
\usepackage{longtable} 
\usepackage{graphicx}
\usepackage{subcaption}
\usepackage{calc}
\usepackage[most]{tcolorbox}
\usepackage{listings}
\usepackage{wrapfig}
\usepackage{caption}
\usepackage{graphicx}

\usepackage[utf8]{inputenc} 
\usepackage[T1]{fontenc}    
\usepackage{hyperref}       
\usepackage{url}            
\usepackage{booktabs}       
\usepackage{amsfonts}       
\usepackage{nicefrac}       
\usepackage{microtype}      
\usepackage{xcolor}         
\usepackage{graphicx}
\usepackage{subcaption}
\usepackage{bm}
\usepackage{makecell}

\definecolor{uclablue}{rgb}{0.15, 0.45, 0.68}
\hypersetup{
    breaklinks,
    citecolor=uclablue,
    colorlinks=true,
}

\newtcolorbox{AIbox}[2][]{aibox,title=#2,#1}
\tcbset{
  aibox/.style={
    top=10pt,
    colframe=black,
    colbacktitle=black,
    coltitle=white,
    center,
  }
}

\lstdefinelanguage{prompt}{
    basicstyle=\scriptsize\ttfamily, 
    mathescape=true,        
    escapebegin=\color{latentcolor},  
    escapeend={},
    escapechar=@,
    stringstyle = \color{myorange},
    showstringspaces = false,
    moredelim = [s][\color{mypink}]{`}{`},
    moredelim = [s][\color{mybrown}]{```json}{```},
    moredelim = [s][\color{latentcolor}]{<StartOfLatent>}{<EndOfLatent>},
    literate = %
        {\ \ a.\ }{{\textcolor{mypurple}{\ \ a.\ }}}5
        {\ \ b.\ }{{\textcolor{mypurple}{\ \ b.\ }}}5
        {\ \ c.\ }{{\textcolor{mypurple}{\ \ c.\ }}}5
        {\ \ d.\ }{{\textcolor{mypurple}{\ \ d.\ }}}5
        {\ \ e.\ }{{\textcolor{mypurple}{\ \ e.\ }}}5
        {\ \ f.\ }{{\textcolor{mypurple}{\ \ f.\ }}}5
        {\ \ g.\ }{{\textcolor{mypurple}{\ \ g.\ }}}5
        {\ \ h.\ }{{\textcolor{mypurple}{\ \ h.\ }}}5
        {\ I.\ }{{\textcolor{mypurple}{\ I.\ }}}4
        {\ II.\ }{{\textcolor{mypurple}{\ II.\ }}}5
        {\ III.\ }{{\textcolor{mypurple}{\ III.\ }}}6
        {\ IV.\ }{{\textcolor{mypurple}{\ IV.\ }}}5
        {\ V.\ }{{\textcolor{mypurple}{\ V.\ }}}4
}

\newtcblisting[auto counter, number within=subsection]{prompt}[4][]{%
  before skip=6pt, after skip=6pt,    
  top=6pt, bottom=6pt, left=7pt, right=7pt, 
  enhanced, breakable,
  colback=white, colframe=gray!65,
  boxrule=0.6pt, arc=2pt,
  drop shadow={black!15},             
  listing only,
  listing options={
    language=prompt,
    upquote=true,
    basicstyle=\scriptsize\ttfamily \setlength{\baselineskip}{1.1\baselineskip},
    columns=fullflexible,
    breaklines=true,
    breakindent=0pt,
    xleftmargin=\ifx\empty#2\empty0pt\else#2\fi,
    xrightmargin=\ifx\empty#3\empty0pt\else#3\fi,
    aboveskip=0pt,                    
    belowskip=0pt,                    
    showstringspaces=false,
  },
  title={\ifstrempty{#4}{Prompt \thetcbcounter}{Prompt \thetcbcounter: #4}},
  colbacktitle=gray!6, coltitle=black,
  attach boxed title to top left={yshift=-1mm, xshift=6pt},
  boxed title style={colback=gray!6, boxrule=0pt, sharp corners},
  #1
}

\newtcblisting[auto counter, number within=subsection]{showcase}[4][]{%
  before=\par\vspace{\baselineskip},
  after=\par,
  width=\linewidth,
  enhanced,
  arc=0em,
  boxrule=1pt,
  listing only,
  listing options={
    language=prompt,
    upquote=true,
    basicstyle=\scriptsize\ttfamily \setlength{\baselineskip}{1.1\baselineskip},
    breaklines=true,
    breakindent=0pt,
    xleftmargin=\ifx\empty#2\empty-12pt\else#2\fi,
    xrightmargin=\ifx\empty#3\empty-5pt\else#3\fi,
    aboveskip=-4pt,
    belowskip=-4pt,
    columns=fullflexible,
    },
  colback=white,
  colframe=gray,
  colbacktitle=gray!5,
  coltitle=black,
  attach boxed title to top center={yshift=-3mm},
  box align=center,
  parbox=false,
  title={\ifstrempty{#4}{Example \thetcbcounter}{Example \thetcbcounter: #4}},
  #1
}

\definecolor{linkColor}{rgb}{0.2,0.4,0.6}
\definecolor{myblue}{HTML}{0379AC}
\definecolor{myred}{HTML}{A50E50}
\definecolor{myorange}{RGB}{238, 133, 74}
\definecolor{latentcolor}{named}{cyan}
\definecolor{normalcolor}{RGB}{0, 0, 0}
\usepackage{marvosym}
\definecolor{lightblue1}{rgb}{0.97, 0.985, 1} 
\definecolor{lightblue2}{rgb}{0.92, 0.965, 1} 
\definecolor{lightblue3}{rgb}{0.84, 0.93, 1}
\definecolor{lightblue4}{rgb}{0.74, 0.87, 1}
\definecolor{lightblue5}{rgb}{0.64, 0.81, 1}
\definecolor{lightblue6}{rgb}{0.54, 0.75, 1}

\definecolor{lightgreen1}{rgb}{0.97, 1.00, 0.97}
\definecolor{lightgreen2}{rgb}{0.92, 0.98, 0.92}
\definecolor{lightgreen3}{rgb}{0.84, 0.95, 0.84}
\definecolor{lightgreen4}{rgb}{0.74, 0.91, 0.74}
\definecolor{lightgreen5}{rgb}{0.64, 0.86, 0.64}
\definecolor{lightgreen6}{rgb}{0.54, 0.81, 0.54}

\definecolor{lightorange1}{rgb}{1.00, 0.98, 0.95}
\definecolor{lightorange2}{rgb}{1.00, 0.95, 0.85}
\definecolor{lightorange3}{rgb}{1.00, 0.90, 0.70}
\definecolor{lightorange4}{rgb}{1.00, 0.85, 0.55}
\definecolor{lightorange5}{rgb}{1.00, 0.80, 0.40}
\definecolor{lightorange6}{rgb}{1.00, 0.75, 0.30}

\definecolor{lightpurple1}{rgb}{0.985, 0.97, 1.00}
\definecolor{lightpurple2}{rgb}{0.96, 0.92, 1.00}
\definecolor{lightpurple3}{rgb}{0.93, 0.84, 1.00}
\definecolor{lightpurple4}{rgb}{0.87, 0.74, 1.00}
\definecolor{lightpurple5}{rgb}{0.81, 0.64, 1.00}
\definecolor{lightpurple6}{rgb}{0.75, 0.54, 1.00}

\definecolor{lightred1}{rgb}{1.00, 0.97, 0.97}
\definecolor{lightred2}{rgb}{1.00, 0.92, 0.92}
\definecolor{lightred3}{rgb}{1.00, 0.84, 0.84}
\definecolor{lightred4}{rgb}{1.00, 0.74, 0.74}
\definecolor{lightred5}{rgb}{1.00, 0.64, 0.64}
\definecolor{lightred6}{rgb}{1.00, 0.54, 0.54}

\definecolor{lightcyan1}{rgb}{0.97, 1.00, 1.00}
\definecolor{lightcyan2}{rgb}{0.92, 0.98, 0.98}
\definecolor{lightcyan3}{rgb}{0.84, 0.95, 0.96}
\definecolor{lightcyan4}{rgb}{0.74, 0.91, 0.94}
\definecolor{lightcyan5}{rgb}{0.64, 0.87, 0.92}
\definecolor{lightcyan6}{rgb}{0.54, 0.83, 0.90}

\usepackage{xcolor,colortbl}
\definecolor{Gray}{gray}{0.85}
\definecolor{LightCyan}{rgb}{0.88,1,1}
\definecolor{greyC}{RGB}{180,180,180}
\definecolor{greyL}{RGB}{235,235,235}
\definecolor{citeColor}{RGB}{0,20,115}
\hypersetup{colorlinks,linkcolor={red},citecolor={citeColor},urlcolor={citeColor}}
\definecolor{shadecolor}{rgb}{0.92,0.92,0.92}

\usepackage{graphicx}
\usepackage{subcaption}
\usepackage{url}

\usepackage{booktabs}
\usepackage{multirow}
\usepackage{cleveref}
\crefname{template}{Template}{Template}

\usepackage{enumitem} 
\usepackage[most]{tcolorbox}
\usepackage{pifont}
\definecolor{rliableblue}{RGB}{0, 102, 204} 

\tcbset{
    myblueboxstyle/.style={
        colback=lightblue3, 
        colframe=lightblue5, 
        coltitle=black, 
        fonttitle=\bfseries, 
        boxrule=1pt, 
        width=\linewidth, 
        left=2pt, 
        right=2pt, 
        top=2pt, 
        bottom=2pt, 
        sharp corners, 
        boxsep=2pt
    }
}

\lstdefinestyle{iclrstyle}{
    language=Python,
    basicstyle=\ttfamily\small,  
    columns=fullflexible,        
    keepspaces=true,             
    showspaces=false,            
    showstringspaces=false,      
    commentstyle=\color{gray}\itshape, 
    keywordstyle=\color{codekw}\bfseries, 
    stringstyle=\color{myorange}, 
    escapechar=|,                
    frame=none,                  
    xleftmargin=1.5em,           
    aboveskip=0.5em,             
    belowskip=0.5em,             
    breaklines=true,             
    breakindent=0pt,
}

\usepackage{graphicx}
\usepackage{booktabs}
\usepackage{pifont}
\usepackage[table]{xcolor}  
\usepackage{wrapfig}
\usepackage{multirow}
\usepackage{float}
\usepackage{pifont}
\usepackage{amssymb}

\usepackage[most]{tcolorbox}
\usepackage{multicol}
\usepackage{tcolorbox}
\usepackage{titlesec}
\definecolor{objblue}{RGB}{3,139,221}  
\definecolor{attrred}{RGB}{255,67,67}    
\definecolor{easygreen}{RGB}{0,156,75}  
\definecolor{middleyellow}{RGB}{242,89,34}  
\definecolor{hardred}{RGB}{216,56,58}
\usepackage{colortbl}
\usepackage{array}
\definecolor{BoxBackground}{RGB}{240, 240, 240} 
\definecolor{BoxFrame}{RGB}{0, 0, 0} 
\definecolor{TitleBackground}{RGB}{0, 0, 0} 
\definecolor{TitleText}{RGB}{255, 255, 255} 
\tcbset{
  academicbox/.style={
    boxsep=5pt,
    left=2pt,
    right=2pt,
    bottom=0.5pt,
    boxrule=0.5pt,
    colback=BoxBackground,
    colframe=BoxFrame,
    colbacktitle=TitleBackground,
    coltitle=TitleText,
    enhanced,
    attach boxed title to top left={yshift=-0.1in,xshift=0.1in},
    boxed title style={boxrule=0pt,colframe=white},
    title={#1},
  }
}
\newtcolorbox{AcademicBox}[1][]{academicbox=#1}

\usepackage{algorithm}
\usepackage{algpseudocode}
\usepackage{listings} 
\usepackage{xcolor}   
\usepackage{etoolbox}
\makeatletter
\AfterEndEnvironment{algorithm}{\let\@algcomment\relax}
\AtEndEnvironment{algorithm}{\kern2pt\hrule\relax\vskip3pt\@algcomment}
\let\@algcomment\relax
\newcommand\algcomment[1]{\def\@algcomment{\footnotesize#1}}
\renewcommand\fs@ruled{\def\@fs@cfont{\bfseries}\let\@fs@capt\floatc@ruled
  \def\@fs@pre{\hrule height.8pt depth0pt \kern2pt}%
  \def\@fs@post{}%
  \def\@fs@mid{\kern2pt\hrule\kern2pt}%
  \let\@fs@iftopcapt\iftrue}
\makeatother

\NewDocumentCommand{\xx}
{ mO{} }{\textcolor{blue}{\textsuperscript{\textit{todo}}\textsf{\textbf{\small[#1]}}}}
\usepackage{xspace}

\usepackage{soul}
\definecolor{codeblue}{rgb}{0.25,0.5,0.5}
\definecolor{codekw}{rgb}{0.85, 0.18, 0.50}
\definecolor{diffgreen}{rgb}{0.0, 0.6, 0.0} 
\definecolor{diffred}{rgb}{0.8, 0.0, 0.0}   

\title{Baton: Explicit Semantic Blueprints for Joint Video-Audio Generation} 

\author{
Shuyuan Tu$^{1,2, *, \ddagger}$, 
Qi Tian$^{2, *}$,  
Zihan Yang$^{1}$, 
Yue Wu$^{2}$, 
Xintong Han$^{2}$, \\
Weijie Kong$^{2}$, 
Jiangfeng Xiong$^{2}$, 
Jian-Wei Zhang$^{2, \dagger}$, \\
Zhao Zhong$^{2}$, 
Liefeng Bo$^{2}$,   
Zuxuan Wu$^{1, \S}$, 
Yu-Gang Jiang$^{1}$\\
\textbf{$^1$Fudan University}  \textbf{$^2$Tencent Hunyuan}\\
$^*$ Equal Contribution, $^\S$ Corresponding Authors, $^\dagger$  Project Leader \\
{\url{https://francis-rings.github.io/Baton}}
}

\begin{document}
\maketitle
\renewcommand\twocolumn[1][]{#1}
\begin{center}
    \centering
    \includegraphics[width=1\textwidth]{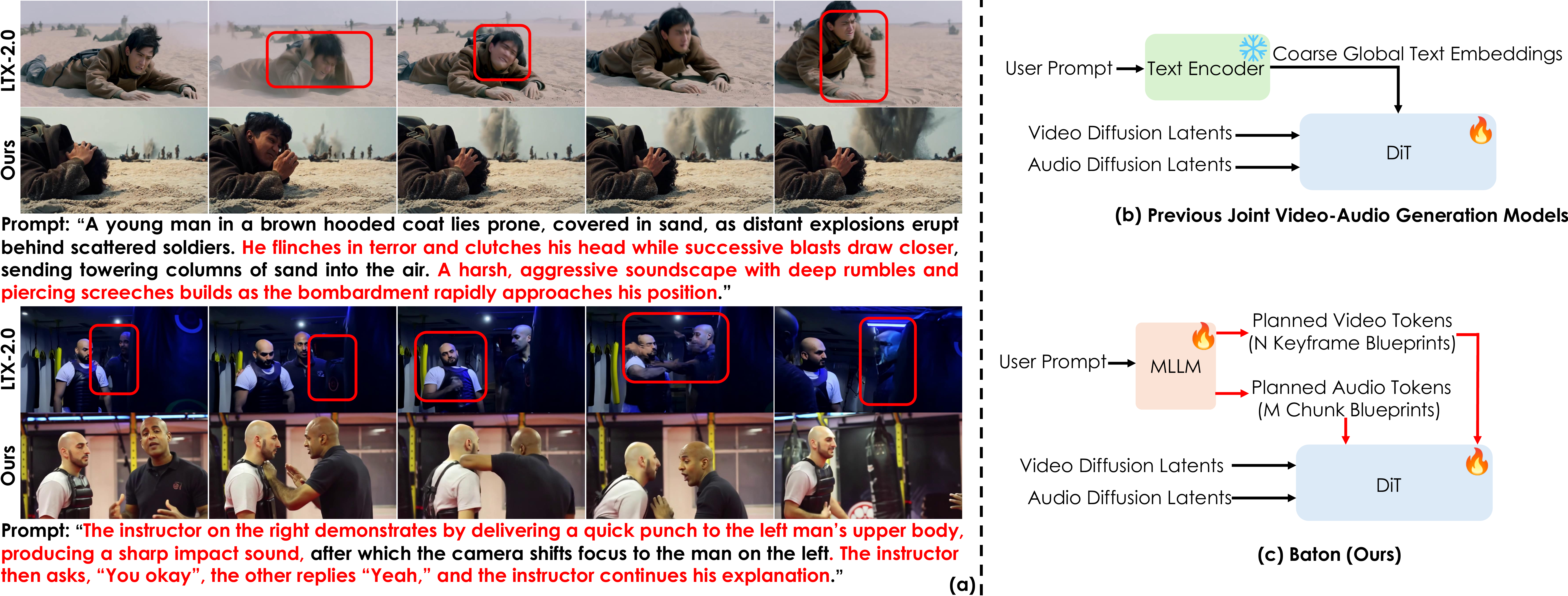}
    \vspace{-0.5cm}
    \captionof{figure}{ (a) Videos generated by Baton, showing its power to synthesize stable video-audio contents in semantically complex scenarios. 
    LTX-2.0 is the latest joint video-audio model. (b) and (c) Framework comparison between previous models and Baton. Please refer to the demo for audio.
    }
    \label{fig:cover}
\end{center}

\renewcommand*{\thefootnote}{\fnsymbol{footnote}}
\footnotetext{$^\ddagger$ Work done when interning at Tencent Hunyuan.}

\begin{abstract}
  Current open-source diffusion models struggle to generate stable and synchronized audio-visual content, particularly in scenarios demanding complex semantic reasoning. The root cause is that existing methods rely on coarse text embeddings from off-the-shelf encoders to guide audio-video denoising, which discards fine-grained semantics and, critically, lacks a shared long-horizon plan, leading to uncoordinated denoising trajectories and fragile cross-modal alignment.
We propose Baton, the first framework that introduces explicit semantic planning into joint video-audio generation. Our key insight is that complementing coarse text guidance with semantically rich, modality-aware planned tokens, jointly reasoned and mutually aligned before denoising, can simultaneously restore fine-grained semantic detail and establish a shared blueprint that coordinates both audio and video denoising trajectories. 
Concretely, Baton first introduces the VA-Planner, a multimodal language model equipped with dual semantic alignment towers, where learnable queries cross-attend to both video and audio features to produce a pair of semantically aligned video and audio planned tokens as keyframe-level blueprints. 
These planned tokens are injected into the diffusion backbone via cross-attention layers, providing temporally grounded guidance complementary to coarse text embeddings. Since planned tokens do not share one-to-one spatial-temporal correspondence with diffusion latents, we further propose Relative Semantic RoPE, a relative positional encoding that maps planned tokens and latents into a shared spatial-temporal coordinate frame, enabling each latent to accurately attend to its positionally corresponding semantic cues. Experiments on benchmarks show the effectiveness of Baton both qualitatively and quantitatively.
\end{abstract}
\section{Introduction}

Previous video generation models~\citep{bao2024vidu, hong2022cogvideo, kong2024hunyuanvideo, wan2025} lack audio, limiting their real-world applicability. It drives joint video–audio generation to the mainstream in generative AI, as natively synchronized outputs are far more natural than video-first pipelines with post-hoc dubbing~\citep{hu2025harmony, zhang2025uniavgen}.
However, current open-source models~\citep{zhang2025uniavgen,low2025ovi,hacohen2026ltx,team2026mova} struggle to handle scenarios demanding complex semantic reasoning, such as compositional instructions with multi-stage actions or human–object interactions. It requires reasoning over long-range causal relationships, where generating stable video-audio remains significantly challenging.

While current open-source models~\citep{liu2025javisdit, liu2025javisgpt, low2025ovi, hacohen2026ltx, ruan2023mm, wang2025universe, hu2025harmony, zhang2025uniavgen, team2026mova} explore various strategies to enhance video-audio synchronization, they focus on simple prompts and use coarse text embeddings from off-the-shelf encoders to guide audio-video denoising, yet none of them design a module for cross-modal semantic comprehension that models how actions and causal intentions should coherently manifest and temporally progress across the visual and auditory streams (Fig. \ref{fig:cover} (b/c)). 
Such global embeddings lack fine-grained understanding of how each event should jointly manifest across both modalities in a temporally coherent manner, causing the two denoising trajectories to drift apart and become adversarial, leading to severe cross-modal misalignment or distortion.
Thus, these methods struggle to synthesize stable video-audio when confronted with semantically complex prompts. 
While some works~\citep{team2026mova,hacohen2026ltx} apply LLMs~\citep{yang2025qwen3, Qwen2.5-VL, Qwen3-VL} to enrich prompt details, they remain global embeddings injected into diffusion models~\citep{dhariwal2021diffusion, ho2020denoising, ho2022cascaded, song2020score, song2020denoising, rombach2022high}, which still cannot tackle the above issues.

In light of this, we propose Baton to perform semantic planning for stable joint video-audio generation, which explicitly disentangles semantic reasoning and synthesis (Fig. \ref{fig:framework}).
In particular, Baton first introduces the VA-Planner, which is a learnable multimodal language model (MLLM) with dual semantic alignment towers to conduct semantic reasoning. 
Each tower uses learnable queries that cross-attend to both video and audio features, acting as modality bridges to fuse semantic cues and capture cross-modal correspondence.
Unlike global text embeddings, VA-Planner reasons over the user prompts to produce a pair of semantically aligned video and audio planned tokens. 
These tokens, coordinated via the above towers, serve as keyframe-level blueprints that encode how visual scenes and their accompanying sounds should jointly unfold over time, providing fine-grained, modality-aware guidance that global text embeddings cannot offer. 
The planned tokens are then fed into the diffusion model via cross-attention, grounding each denoising step with temporally structured semantic cues. 
As cross-modal semantic coordination is explicitly established before denoising, both the video and audio generation trajectories are anchored to a shared, pre-aligned semantic roadmap. 
It prevents the two modalities from drifting into adversarial dynamics, as both trajectories are guided to evolve along semantically coherent paths rather than blindly following vague signals.

However, since planned tokens and diffusion latents do not share one-to-one spatial-temporal correspondence, Baton introduces Relative Semantic RoPE (RS-RoPE), a relative positional encoding that maps them into a shared coordinate frame so that each latent can accurately attend to its positionally corresponding semantic guidance. Thus, Baton transforms the generation from blindly denoising under vague global signals to following a semantically grounded and cross-modally coordinated plan, yielding stable video-audio content even for prompts requiring semantic reasoning.

As shown in Fig. \ref{fig:cover} (a), while LTX-2~\citep{hacohen2026ltx} suffers from body distortion and misalignment between audio and video, Baton can accurately synthesize stable video-audio content based on user prompts, even in long-horizon scenarios involving multiple actions and complex human–object interactions. 

In conclusion, our contributions are as follows: 
(1) We propose the novel VA-Planner to perform semantic reasoning over user prompts to produce a pair of semantically aligned video and audio planned tokens as keyframe-level blueprints, anchoring both denoising trajectories to a shared, coordinated semantic plan.
To our knowledge, Baton is the first to disentangle semantic planning with synthesis for joint video-audio generation. 
(2) We introduce the RS-RoPE, which maps planned tokens and diffusion latents into a shared spatial-temporal coordination, enabling each latent to attend to its corresponding semantic cues.
(3) Experiments across benchmarks show the superiority of Baton over the SOTA. 

\section{Related Work}

\noindent\textbf{Video Generation.}
The capacity of high fidelity and diversity in diffusion models~\citep{dhariwal2021diffusion,ho2020denoising,ho2022cascaded, tu2023implicit,nichol2021improved,song2020score,song2020denoising,rombach2022high,meng2021sdedit,hertz2022prompt,tumanyan2023plug, tu2024motioneditor, tu2024motionfollower, tu2025stableanimator, tu2025stableanimator++} has promoted dramatic interest in their application for video generation~\citep{cui2025hallo3, ji2025sonic, kong2025let, gan2025omniavatar, xu2024hallo, yang2025infinitetalk, tu2025stableanimator++,tu2025stableavatar,tu2025flashportrait,huang2025plan}.
Early video diffusion models~\citep{singer2022make, blattmann2023stable, guo2023animatediff, wu2023tune, wang2024magicvideo, videoworldsimulators2024,tu2024motioneditor,tu2024motionfollower,tu2025stableanimator} mostly utilize the U-Net for video generation by inserting temporal layers into the backbone.
Recent works~\citep{bao2024vidu, hong2022cogvideo, kong2024hunyuanvideo, wan2025, tu2025stableavatar, tu2025flashportrait, yang2026arcflow} replace the U-Net with the Diffusion-in-Transformer (DiT)~\citep{peebles2023scalable} for scalability.
However, the above models only generate silent videos, restricting their practical applications.

\noindent\textbf{Joint Video-Audio Generation.}
The high fidelity of industry-leading models (Veo 3~\citep{veo3_2025} and Seedance 2.0~\citep{seedance2_2025}) sparks the research interest in joint video-audio generation~\citep{haji2025av, ishii2025simple, liu2025javisdit, liu2025javisgpt, low2025ovi, hacohen2026ltx, ruan2023mm, wang2025universe, hu2025harmony, zhang2025uniavgen, team2026mova}. JavisDiT~\citep{liu2025javisdit} and UniAVGen~\citep{zhang2025uniavgen} focus on joint video-speech generation. UniVerse-1~\citep{wang2025universe} and Ovi~\citep{low2025ovi} can also synthesize ambient sounds.
Harmony~\citep{hu2025harmony} enhances the synchronization via cross-task synergy training. MOVA~\citep{team2026mova} and LTX-2~\citep{hacohen2026ltx} improve the quality by scaling the model and data.
JavisGPT~\citep{liu2025javisgpt} utilizes unified tokens to guide the generation.
However, the above models mostly apply frozen language models (T5~\citep{raffel2020exploring} or Qwen~\citep{yang2025qwen3}) to extract the global text embeddings and inject them into DiT. 
Thus, they struggle to synthesize stable audio-visual content in semantically complex scenarios.
By contrast, Baton addresses these issues.

\noindent\textbf{Unified MultiModal Generation.}
Some works~\citep{chen2025blip3, deng2025emerging, han2025vision, kim2025democratizing, liao2025mogao, lin2025bifrost, ma2025janusflow, qu2025tokenflow, sun2024generative, wang2025selftok, wu2024vila, xie2024show, xie2025x, yao2025reconstruction, zheng2025diffusion, zhou2024transfusion, tu2023implicit} tend to achieve multimodal understanding and generation via a single framework.
Early works~\citep{kim2025democratizing, wang2025selftok, wang2024emu3, yao2025reconstruction, han2025vision} convert pretrained semantic encoders~\citep{tschannen2025siglip} to text-aligned visual encoders for generation. Recent image generation models~\citep{chen2025blip3o, ge2024seed, deng2025emerging, xie2024show, zhou2024transfusion, chen2025blip3, pan2025transfer} use MLLMs to predict visual tokens, coupled with diffusion heads for image synthesis.
However, these methods primarily focus on static image generation, with limited exploration of joint video–audio generation under temporal semantic reasoning, such as multi-stage actions or complex human–object interactions.

\section{Method}

Shown in Fig. \ref{fig:framework}, Baton explicitly disentangles semantic reasoning and synthesis for establishing a modality-aware blueprint that coordinates both audio and video
denoising. 
Concretely, the prompts are first fed to an MLLM to perform semantic reasoning and predict a pair of video/audio planned tokens as shared blueprints, as detailed in Sec. \ref{sec: va_planner}. 
To provide blueprint guidance, the planned tokens are further injected into a DiT via cross-attention. The DiT has the same dual-branch architecture as Ovi~\citep{low2025ovi}. 
Notably, as the planned tokens and the diffusion latents live on misaligned spatio-temporal grids, we introduce Relative Semantic RoPE to resolve the mismatch, as detailed in Sec. \ref{sec: relative_semantic_rope}.

\begin{figure*}[t!]
\begin{center}
\includegraphics[width=0.95\linewidth]{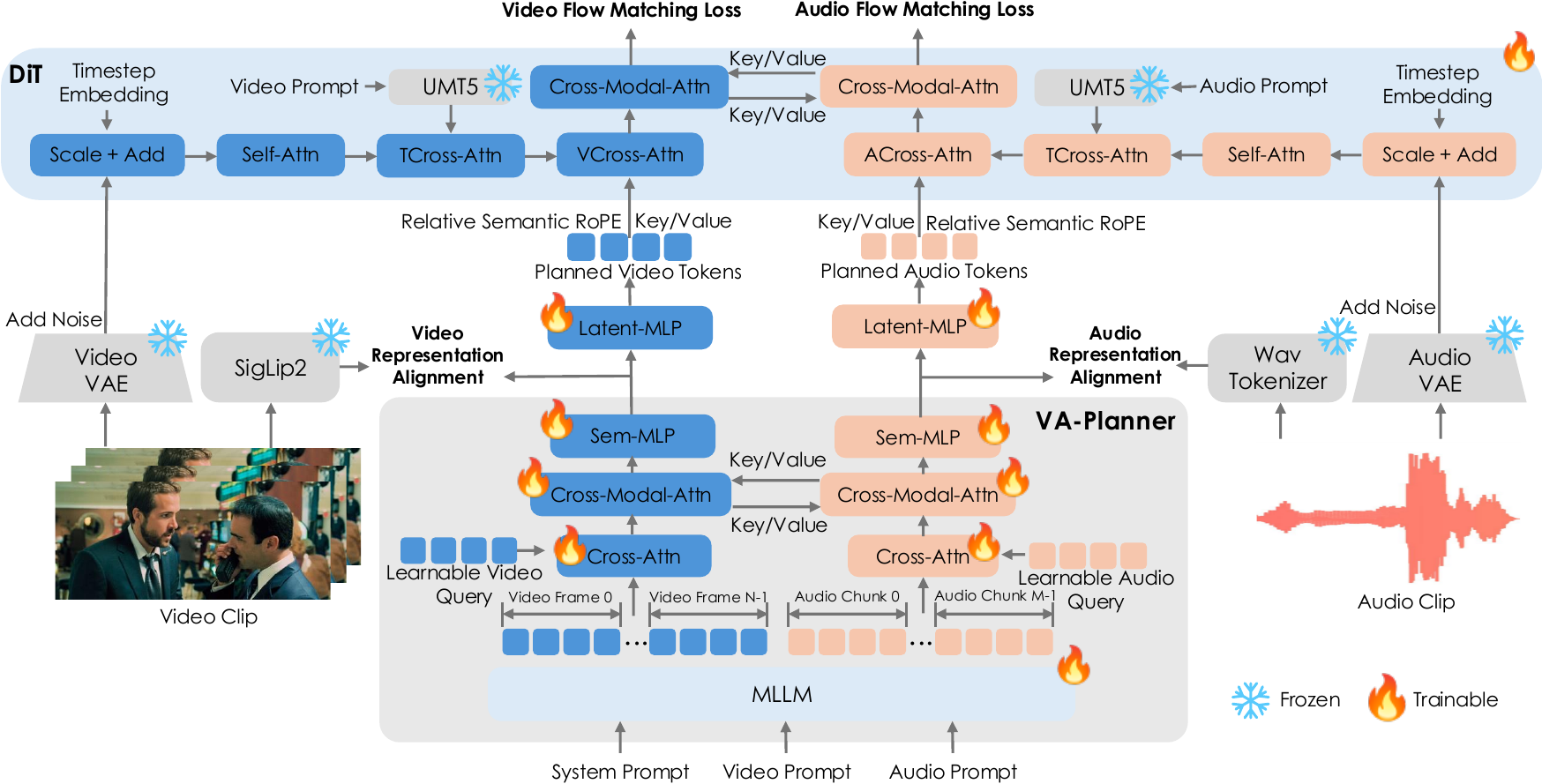}
\end{center}
\vspace{-0.3cm}
   \caption{Architecture of Baton. 
   Given the user prompts, Baton utilizes an MLLM to perform semantic reasoning and predict planned video/audio tokens, which are fed to the DiT via cross-attention for offering keyframe-level detailed blueprints during joint video-audio denoising.
   }
\label{fig:framework}
\vspace{-0.3cm}
\end{figure*}

\subsection{VA-Planner}
\label{sec: va_planner}

Current joint video-audio generation models~\citep{hacohen2026ltx,team2026mova,zhang2025uniavgen,liu2025javisdit,hu2025harmony} only rely on global text embeddings gained from a frozen LLM~\citep{raffel2020exploring, yang2025qwen3} to guide the generation. It encodes the entire prompt into a vague embedding without decomposing it into modality-specific temporal semantics or modeling how visual events and auditory cues should correspond at each generation stage, leaving both denoising branches to independently interpret the vague signal and inevitably diverge under complex scenarios.
To tackle this, we propose the VA-Planner, which uses a trainable MLLM to perform semantic reasoning and predict modality-specific yet mutually aligned planned tokens. 
Each token encodes a localized semantic context specifying what occurs, where, and when, while the paired video and audio tokens are jointly reasoned in a single autoregressive pass to ensure cross-modal consistency at every time point. 
Both generation trajectories are anchored to a shared semantic roadmap before denoising, preventing the two modalities from drifting into adversarial dynamics.
It can restore fine-grained semantics and establish the shared cross-modal blueprints that global embeddings cannot provide.

Concretely, to plan a video with $N$ keyframes (sampled at $\text{FPS}=6$) and $M$ audio chunks (each chunk covers one second of audio), we construct a structured user prompt $T_{user}$ (Appx. \ref{sec: training_details}) that concatenates the system prompt $T_{sys}$, video prompt $T_v$, audio prompt $T_a$, and special delimiter tokens ($T^{tag}_{v}$ and $T^{tag}_{a}$) that delineate the visual and auditory planning regions:
\begin{equation}\small \label{eq:user_prompts}
\begin{aligned}
T_{\text{user}} &= [T_{\text{sys}};\; T_v;\; T_a;\; T^{tag}_{v};\; T^{tag}_{a}],
\end{aligned}
\end{equation}
where $T^{tag}_{v}=\texttt{<|v\_start|>}; \texttt{<|img\_pad|>} \cdots \texttt{<|img\_pad|>}; \texttt{<|v\_end|>}$. Each \texttt{<|img\_pad|>} is a placeholder for one visual semantic token, with $n_v$ tokens per keyframe, yielding $L_v = N \times n_v$ video tokens in total. 
Since predicting visual tokens for all frames incurs prohibitive cost, we sample $N$ keyframes at $\text{FPS}=6$ instead of all video frames.
$T^{tag}_{a}=\texttt{<|a\_start|>}; \texttt{<|aud\_pad|>} \cdots \texttt{<|aud\_pad|>}; \texttt{<|a\_end|>}$.
Each \texttt{<|aud\_pad|>} is a placeholder for one audio semantic token, with $n_a$ tokens per chunk, yielding $L_a = M \times n_a$ audio tokens. 
Since $n_a << n_v$, predicting all audio chunks is affordable. 
We adopt the same multimodal RoPE as Qwen3-VL~\citep{Qwen3-VL} in the MLLM to jointly encode the heterogeneous positions of text, video, and audio tokens in $T_{user}$.
To encode how the described visual scenes and accompanying sounds should semantically unfold over time, the MLLM performs autoregressive reasoning over $T_{user}$, and we extract the hidden states at the pad token positions to gain the video and audio hidden states:
\begin{equation}\small
\label{eq:mllm_hidden_state}
\begin{aligned}
     \bm{H}_v = \mathtt{MLLM}(T_{user})_{[\texttt{v\_start+1}:\texttt{v\_end}]} \in \mathbb{R}^{L_v \times D}, \quad \bm{H}_a = \mathtt{MLLM}(T_{user})_{[\texttt{a\_start+1}:\texttt{a\_end}]} \in \mathbb{R}^{L_a \times D},
\end{aligned}
\end{equation}
where $D$ is the hidden dimension. 
By placing the video and audio planning regions after all text prompts within a single autoregressive sequence, the causal attention of MLLM naturally allows both $\bm{H}_v$ and $\bm{H}_a$ to condition on the full prompt context, while $\bm{H}_a$ further attends to the preceding $\bm{H}_v$, establishing an implicit cross-modal dependency at the reasoning stage itself.

Since the planned tokens aim to encode concrete perceptual structures rather than remain in the MLLM's language-centric space, we introduce dual semantic alignment towers that project them into the continuous feature domains of pretrained perceptual encoders (SigLip2~\citep{tschannen2025siglip} for video and WavTokenizer~\citep{ji2024wavtokenizer} for audio). 
As the MLLM's causal dependency is unidirectional ($\bm{H}_v$ cannot attend to $\bm{H}_a$), the video plan is blind to audio cues. 
The towers address this via bidirectional cross-modal attention. 
Each tower employs learnable queries, allowing them to flexibly distill the most relevant semantic cues from $\bm{H}_v$ and $\bm{H}_a$ into planned tokens.
For the video tower, the learnable query $\bm{Q}_v \in \mathbb{R}^{L_v \times D}$ first cross-attends to $\bm{H}_v$ to extract video-specific semantics, then performs cross-modal attention $\mathtt{CMAttn}_{a \rightarrow v}(\cdot)$ with $\bm{H}_a$ to absorb complementary auditory cues, and is projected to the target perceptual encoder dimension (SigLip2 or WavTokenizer) via a Sem-MLP $\mathtt{SMLP}_{v/a}(\cdot)$:
\begin{equation}\small
\label{eq:video_tower_mllm}
\begin{aligned}
     \bm{H}^{sem}_v = \mathtt{SMLP}_{v}(\mathtt{CMAttn}_{a \rightarrow v}(\mathtt{CAttn}_{v}(\bm{Q}_v, \bm{H}_v), \bm{H}_a)),
\end{aligned}
\end{equation}
where $\mathtt{CAttn}_{v}(x, y)$ and $\mathtt{CMAttn}_{a \rightarrow v}(x, y)$ refer to cross-attention with $x$ as Query and $y$ as Key/Value. 
Symmetrically, the audio tower produces:
\begin{equation}\small
\label{eq:audio_tower_mllm}
\begin{aligned}
     \bm{H}^{sem}_a = \mathtt{SMLP}_{a}(\mathtt{CMAttn}_{v \rightarrow a}(\mathtt{CAttn}_{a}(\bm{Q}_a, \bm{H}_a), \bm{H}_v)),
\end{aligned}
\end{equation}
where $\bm{Q}_a$ is the learnable query. 
Since $\bm{H}_v$ and $\bm{H}_a$ have different temporal densities, we apply timestamp-based RoPE in $\mathtt{CMAttn}_{v \rightarrow a}(\cdot)$ and $\mathtt{CMAttn}_{a \rightarrow v}(\cdot)$ to map both modalities onto a shared time axis, as detailed in Appx. \ref{sec: dual_semantic_alignment_tower_details}. 
Due to the towers, $\bm{H}^{sem}_v$ and $\bm{H}^{sem}_a$ both encode a mutually consistent temporal blueprint rather than two independent plans.
Notably, Baton employs two distinct RoPE designs at different stages: (i) timestamp-based RoPE within the dual towers' $\mathtt{CMAttn}$ for aligning cross-modal tokens during planning (Appx.~\ref{sec: dual_semantic_alignment_tower_details}), and (ii) Relative Semantic RoPE in the DiT's $\mathtt{VCAttn}$/$\mathtt{ACAttn}$ for aligning planned tokens with diffusion latents during denoising (Sec.~\ref{sec: relative_semantic_rope}).

\subsection{Relative Semantic RoPE}
\label{sec: relative_semantic_rope}

To provide keyframe-level blueprints during denoising, we inject $\bm{H}^{sem}_v$ and $\bm{H}^{sem}_a$ into dual-branch-based DiT via cross attention. 
However, the video/audio diffusion latents and $\bm{H}^{sem}_v$/$\bm{H}^{sem}_a$ do not share one-to-one spatiotemporal correspondence (Fig. \ref{fig:rope_motivation}) due to mismatched compression rates. Latents attend uniformly to all planned tokens by content similarity alone, collapsing the structured plan into a vague global signal. To address this, we propose the Relative Semantic RoPE (RS-RoPE).

In particular, we first employ Latent-MLP $\mathtt{LMLP}_{v/a}(\cdot)$ to project planned tokens $\bm{H}^{sem}_v$ and $\bm{H}^{sem}_a$ from the spaces of SigLip2 and WavTokenizer to the diffusion latent dimension $d$:
\begin{equation}\small
\label{eq:latent_mlp}
\begin{aligned}
     \bm{H}^{dit}_v = \mathtt{LMLP}_v(\bm{H}^{sem}_v) \in \mathbb{R}^{L_v \times d}, \quad \bm{H}^{dit}_a = \mathtt{LMLP}_a(\bm{H}^{sem}_a) \in \mathbb{R}^{L_a \times d},
\end{aligned}
\end{equation}
$\bm{H}^{dit}_v$ and $\bm{H}^{dit}_a$ subsequently perform VCross-Attn $\mathtt{VCAttn}(\cdot)$ and ACross-Attn $\mathtt{ACAttn}(\cdot)$ with diffusion latents. However, the queries and keys occupy different grids. 
For example, The video diffusion latents $\bm{z}_v$, compressed by the 3D VAE, live on a $(T_v^l, H_v^l, W_v^l)$ grid, while $\bm{H}^{dit}_v$ are arranged on a $(N, h_s, w_s)$ semantic grid ($n_v = h_s \times w_s$ spatial tokens per keyframe). The two grids differ along all three axes, so tokens representing the same space-time point get different indices under standard RoPE, leading to misaligned encoding and incorrect attention.

Thus, we introduce the Relative Semantic RoPE, which rescales the position indices of both grids into a shared coordinate system.
Regarding $\mathtt{VCAttn}(\cdot)$, given a latent query at grid index $(i_t, i_h, i_w)$ and a planned semantic key at $(j_t, j_h, j_w)$, the query and key position indices are $\bm{p}_{v}^q = (i_t, \; i_h, \; i_w)$ and $\bm{p}_{v}^k = \!\left(j_t \!\cdot\! \frac{T_v^l}{N}, \;\; j_h \!\cdot\! \frac{H_v^l}{h_s}, \;\; j_w \!\cdot\! \frac{W_v^l}{w_s}\right)$.
We then inject $\bm{H}^{dit}_v$ to the DiT:
\begin{equation}\small
\label{eq:vcattention}
\begin{aligned}
     \mathtt{RoPE}(\bm{h}_*, p) &= \bm{h}_* \odot \cos(p \cdot \boldsymbol{\omega}) + \mathtt{Rotate}(\bm{h}_*) \odot \sin(p \cdot \boldsymbol{\omega}), \\
     \mathtt{RoPE}_{3D}(\bm{h}, \bm{p}) &= \big[\mathtt{RoPE}(\bm{h}^{(t)}\!,\, p_t);\;\; \mathtt{RoPE}(\bm{h}^{(h)}\!,\, p_h);\;\; \mathtt{RoPE}(\bm{h}^{(w)}\!,\, p_w)\big], \\
     \mathtt{VCAttn}(\bm{z}_v, \bm{H}^{dit}_v)&=\mathtt{Attn}(\mathtt{RoPE}_{3D}(\bm{z}_v, \bm{p}_{v}^q), \mathtt{RoPE}_{3D}(\bm{H}^{dit}_v, \bm{p}_{v}^k), \bm{H}^{dit}_v),
\end{aligned}
\end{equation}
where $\mathtt{Attn}(\cdot)$ is the attention operation.
$\mathtt{RoPE}_{3D}(\cdot)$ is applied per attention head. For each head, the query or key vector $\bm{h}\in \mathbb{R}^{d_h}$ is partitioned into $[\bm{h}^{(t)};\, \bm{h}^{(h)};\, \bm{h}^{(w)}]$ corresponding to the temporal, height, and width axes. The frequency vector $\omega_k = \frac{1}{\theta^{2k/d_h}}$ ($\theta$ is a hyperparameter) and $\mathtt{Rotate}(\cdot)$ swaps adjacent pairs with sign flips following the standard RoPE~\citep{su2024roformer}.
As our RoPE encodes only relative position, it endows the cross-attention with a soft spatio-temporal locality prior that preserves the keyframe-level structure of the blueprints, rather than flattening it into a global context.

Symmetrically, $\mathtt{ACAttn}(\cdot)$ uses $\bm{z}_a\in \mathbb{R}^{T_a^l \times d}$ as queries and $\bm{H}^{dit}_a$ as keys and values. Since audio carries no spatial structure, the alignment reduces to 1D RoPE:
\begin{equation}\small
\label{eq:acattention}
\begin{aligned}
\mathtt{ACAttn}(\bm{z}_a, \bm{H}^{dit}_a)=\mathtt{Attn}(\mathtt{RoPE}(\bm{z}_a, p^q_a), \mathtt{RoPE}(\bm{H}^{dit}_a, p^k_a), \bm{H}^{dit}_a),
\end{aligned}
\end{equation}
where $p^q_a(i) = i$ for the $i$-th audio latent and $p^k_a(k) = k \cdot T_a^l / L_a$ for the $k$-th planned audio token, aligning both onto the same temporal axis.
Notably, previous cross-modal RoPE strategies in joint video-audio generation~\citep{low2025ovi,team2026mova,zhang2025uniavgen,hu2025harmony} only handle 1D temporal scaling between homogeneous video and audio latent streams that share the same backbone architecture. Our RoPE goes further by operating in full 3D, simultaneously resolving temporal and spatial mismatches between two fundamentally heterogeneous features (planned semantic tokens and diffusion latents).

\subsection{Training}
\label{sec: training}

Baton is trained in three stages:
(1) \textbf{Stage 1 (VA-Planner Pretraining).}
The VA-Planner aims to autoregressively reason over user prompts and predict planned tokens that capture the perceptual structure of the target content. 
We initialize the MLLM from Qwen3~\citep{yang2025qwen3} and train the entire VA-Planner ($\mathtt{MLLM}$+$\mathtt{SMLP}_{v/a}(\cdot)$). Given ground-truth video and audio, we extract target continuous features $\bm{F}^{gt}_v \in \mathbb{R}^{L_v \times D_s}$ and $\bm{F}^{gt}_a \in \mathbb{R}^{L_a \times D_a}$ from the penultimate layers~\citep{ma2024exploring} of frozen SigLip2 and WavTokenizer. The VA-Planner is supervised as follows:
\begin{equation}\small
\label{eq:loss_plan}
\begin{aligned}
\mathcal{L}_{plan} = \sum_{t=1}^{N}\sum_{i=1}^{n_v} \|\bm{H}^{sem}_{v,(t,i)} - \bm{F}^{gt}_{v,(t,i)}\|_2^2 + \sum_{m=1}^{M}\sum_{j=1}^{n_a} \|\bm{H}^{sem}_{a,(m,j)} - \bm{F}^{gt}_{a,(m,j)}\|_2^2,
\end{aligned}
\end{equation}
where $\bm{H}^{sem}_{v,(t,i)}$ is the planned video token at the $i$-th spatial position of the $t$-th keyframe and $\bm{H}^{sem}_{a,(m,j)}$ is the planned audio token at the $j$-th position of the $m$-th audio chunk. 
By contrast to planning over discrete tokens, regressing continuous features preserves richer semantic structure. 
(2) \textbf{Stage 2 (DiT Adaptation).}
To allow the DiT to learn the semantic feature distribution without being confounded by planner prediction noise, we initialize the DiT from Ovi~\citep{low2025ovi} and feed the ground-truth $\bm{F}^{gt}_v$, $\bm{F}^{gt}_a$ (projected via Latent-MLPs) directly into $\mathtt{VCAttn}(\cdot)$ and $\mathtt{ACAttn}(\cdot)$. 
We use the flow matching loss~\citep{lipman2022flow} to train the DiT $\hat{\bm{v}}_\theta^{v/a}(\cdot)$ (including $\mathtt{LMLP}_{v/a}(\cdot)$):
\begin{equation}\small
\label{eq:loss_fm}
\begin{aligned}
\mathcal{L}_{FM} = \mathbb{E}_{\bm{z}_0^{v/a},\, \bm{z}_1^{v/a},\, t} \big[ \|\hat{\bm{v}}_\theta^v(\bm{z}_t^v, \bm{z}_t^a, t, \bm{c}, \bm{F}^{gt}_v) - (\bm{z}_1^v \!-\! \bm{z}_0^v)\|_2^2 + \|\hat{\bm{v}}_\theta^a(\bm{z}_t^v, \bm{z}_t^a, t, \bm{c}, \bm{F}^{gt}_a) - (\bm{z}_1^a \!-\! \bm{z}_0^a)\|_2^2 \big],
\end{aligned}
\end{equation}
where $\bm{z}_1^{v/a} \sim \mathcal{N}(0, I)$. $\bm{z}_0^{v/a}$ and $\bm{c}$ are the VAE-encoded clean latents and the conditioning signals.
(3) \textbf{Stage 3 (Joint Fine-tuning).}
The VA-Planner and DiT are connected. The VA-Planner is frozen, and the DiT is trainable. 
The DiT now receives $\bm{H}^{sem}_v$ and $\bm{H}^{sem}_a$ as conditioning, and training continues with Eq.\ref{eq:loss_fm}. This stage bridges the gap between the clean encoder features in Stage 2 and the imperfect planner predictions, mitigating exposure bias and ensuring robust generation.
\begin{table}[t!]\small
\caption{Quantitative comparisons with previous open-source methods on Verse-Bench and Sem100. In the table elements a/b, a and b refer to the results on Verse-Bench and Sem100. 
}
\vspace{-0.10in}
\begin{center}
\renewcommand\arraystretch{1.1}
\scalebox{0.68}{
\begin{tabular}{l|ccccccccccc}
\toprule
Model    & AQ$\uparrow$        & IQ$\uparrow$        & DD$\uparrow$        & ID$\uparrow$        & PQ$\uparrow$        & CU$\uparrow$        & M-WER$\downarrow$      & Sync-C$\uparrow$    & Sync-D$\downarrow$      & DeSync$\downarrow$    & P-Acc$\uparrow$     \\ \midrule
Ovi~\citep{low2025ovi}      & 0.56/0.42 & 0.67/0.65 & 0.49/0.36 & 0.90/0.88 & 6.30/6.74 & 5.91/6.23 & 0.75/0.66  & 3.34/5.07 & 8.78/8.25   & 0.54/1.06 & 0.74/0.46 \\
JavisGPT~\citep{liu2025javisgpt} & 0.34/0.28 & 0.45/0.38 & 0.30/0.32 & 0.32/0.45 & 5.37/5.12 & 4.77/5.06 & 4.76/3.96  & 0.65/0.69 & 11.27/12.34 & 1.16/1.18 & 0.43/0.25 \\
UniAVGen~\citep{zhang2025uniavgen} & 0.55/0.31 & 0.67/0.58 & 0.47/0.32 & 0.86/0.82 & 4.94/6.68 & 6.11/5.83 & 1.05/0.72  & 1.95/4.80 & 9.57/9.83   & 0.87/1.14 & 0.76/0.44 \\
LTX-2~\citep{hacohen2026ltx}    & 0.53/0.48 & 0.66/0.65 & \textbf{0.71}/0.38 & 0.92/0.91 & 6.54/7.14 & 6.01/6.91 & 0.64/0.58 & 3.77/6.26 & 8.16/7.72   & 0.27/0.97 & 0.85/0.62 \\
MOVA~\citep{team2026mova}     & 0.54/0.44 & 0.64/0.68 & 0.56/0.40 & 0.90/0.87 & 6.71/6.95 & 6.19/6.94 & 1.49/0.86  & 3.52/5.78 & 10.58/8.90  & 0.64/1.05 & 0.79/0.55 \\ \midrule
Ours     & \textbf{0.58}/\textbf{0.54} & \textbf{0.68/0.73} & 0.64/\textbf{0.48} & \textbf{0.93/0.94} & \textbf{6.79/7.53} & \textbf{6.24/7.16} & \textbf{0.18/0.14}  & \textbf{4.26/8.14} & \textbf{7.68/6.85}   & \textbf{0.57/0.68} & \textbf{0.88/0.82} \\ \bottomrule
\end{tabular}
}
\end{center}
\label{table:quantitative_comparison}
\end{table}

\section{Experiments}
\subsection{Implementation Details}

Our training dataset (1.5 million video-audio clips) is aggregated from OpenHuman-Vid~\citep{li2025openhumanvid}, AudioCaps~\citep{kim2019audiocaps}, WavCaps~\citep{mei2024wavcaps}, and videos collected from the internet (Appx. \ref{sec: dataset_details}).
Following previous works~\citep{team2026mova}, we evaluate our model on Verse-Bench~\citep{wang2025universe}. We conduct additional experiments on 100 unseen videos (10 seconds long) with more complex prompts, referred to the Sem100, selected from the internet to assess the semantic reasoning capability of our model. Our DiT and VA-Planner are initialized by Ovi~\citep{low2025ovi} and Qwen3-8B~\citep{yang2025qwen3}. Our model is trained for 10 epochs in 3 training stages, with a batch size of 1 per GPU. The learning rate is $1e-5$ (Appx. \ref{sec: training_details}).

\subsection{Comparison with State-of-the-Art Methods}

\textbf{Quantitative results.}
Following previous works~\citep{hu2025harmony, zhang2025uniavgen}, we utilize AQ~\citep{hu2025harmony}, IQ~\citep{huang2024vbench}, DD~\citep{huang2024vbench}, and ID~\citep{hu2025harmony} to assess the video quality.
We further apply PQ~\citep{tjandra2025meta_audio_aesthetics}, CU~\citep{tjandra2025meta_audio_aesthetics}, M-WER (Multi-Speaker Word Error Rate)~\citep{radford2023robust}, Sync-C~\citep{chung2016out}, Sync-D~\citep{chung2016out}, and DeSync~\citep{iashin2024synchformer} to assess the audio quality and video-audio synchronization. P-Acc evaluates the prompt following accuracy (Appx. \ref{sec: evaluation_metrics}). We perform quantitative comparisons with previous methods~\citep{liu2025javisgpt, low2025ovi, hacohen2026ltx, team2026mova, zhang2025uniavgen} on Verse-Bench~\citep{wang2025universe} and Sem100, as shown in Table \ref{table:quantitative_comparison}.
Compared to the leading competitor LTX-2~\citep{hacohen2026ltx}, Baton achieves comparable results on Verse-Bench, where the text prompts mostly describe simple, single-event scenarios that do not require deep semantic reasoning. The advantage becomes prominent on Sem100, whose prompts involve complex sequential events, intricate human-object interactions, and multi-speaker dialogues that demand strong semantic reasoning. On Sem100, Baton outperforms LTX-2 by 32\% in P-Acc, 76\% in M-WER, and 30\% in DeSync. The M-WER gap is particularly striking, as multi-speaker scenarios require the model to reason about which character speaks what content and when, exactly the kind of localized, temporally grounded semantics that our planned tokens provide, but global text embeddings cannot decompose. The large P-Acc and M-WER gaps confirm that explicit semantic planning is essential for complex prompts where global text embeddings fail to decompose multi-stage actions and multi-speaker dialogues into temporally grounded guidance.

\begin{figure}[t!]
\begin{center}
\includegraphics[width=1\linewidth]{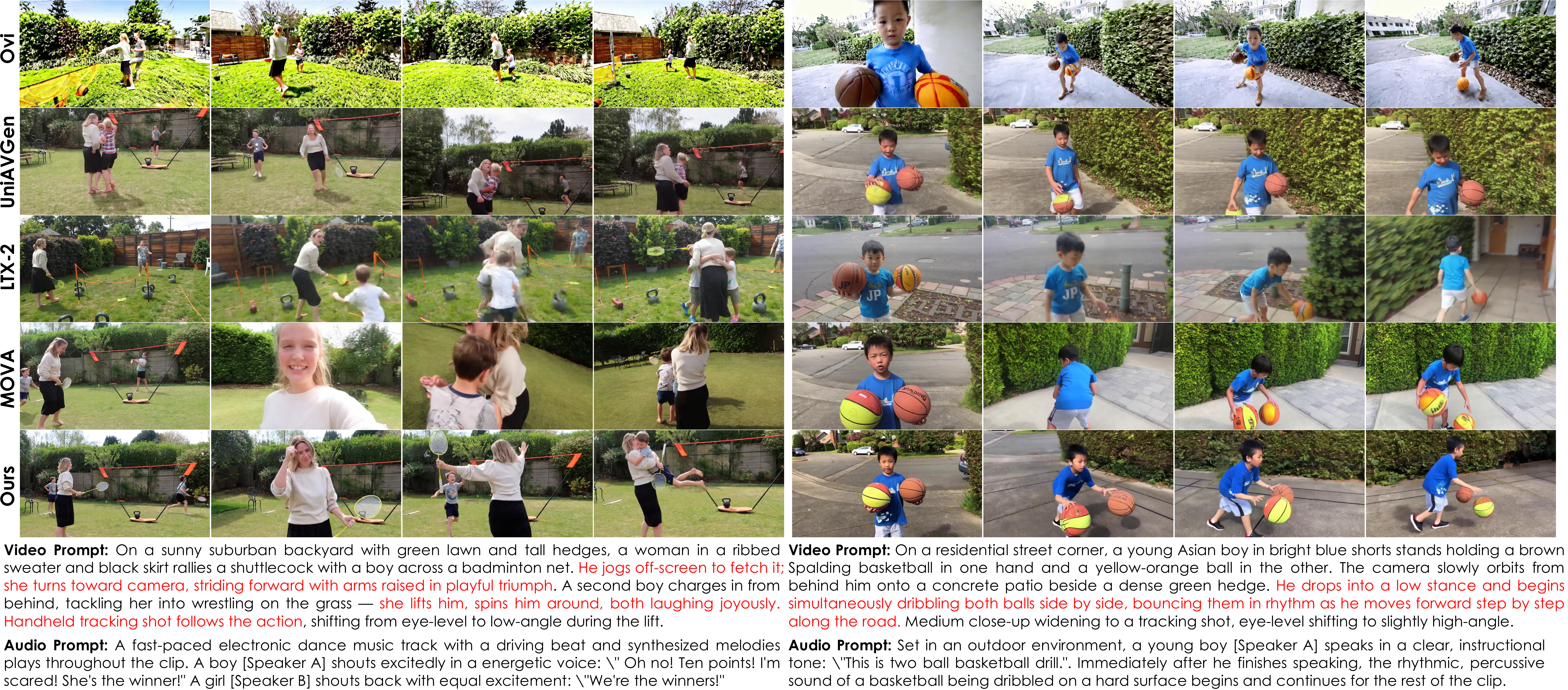}
\end{center}
\vspace{-0.3cm}
   \caption{Qualitative comparisons with previous open-source methods. Please refer to the demo video for audio. More results are in the Appx. \ref{sec: more_comparison_results}.
   }
\label{fig:qualitative_comparison}
\vspace{-0.45cm}
\end{figure}

\noindent\textbf{Qualitative Results.}
The qualitative results are shown in Fig. \ref{fig:qualitative_comparison}.
Notably, since UniAVGen and MOVA only open-source I2VA, we use the first frame of videos generated by Baton as their reference images.
All competitors fail to handle prompts involving sequential events or human-object interactions, indicating that global text embeddings lack structured cross-modal semantic comprehension. Ovi~\citep{low2025ovi} and UniAVGen~\citep{zhang2025uniavgen} suffer from severe body distortion, a result of the two denoising trajectories drifting into adversarial dynamics under complex prompts, where the video and audio branches independently interpret the vague global signal and pull the joint distribution in conflicting directions, ultimately distorting the video. 
LTX-2~\citep{hacohen2026ltx} and MOVA~\citep{team2026mova} exhibit poor subject consistency and audio-video desynchronization. For instance, the basketball in the boy's hands changes color or disappears. 
The global text embeddings cannot model how each object and action should coherently manifest and temporally progress across both modalities, leaving the two trajectories uncoordinated over long horizons. 
In contrast, since Baton establishes cross-modal semantic coordination before denoising, both generation trajectories are anchored to a shared, pre-aligned semantic roadmap, preventing adversarial drift and yielding stable audio-video contents.

\noindent\textbf{Comparison with Commercial Models.}
We compare our model with commercial models~\citep{seedance2_2025,veo3_2025, kling3, wan2_7}, as shown in Appx. \ref{sec: comparison_commercial_models}.
We can see that although commercial models outperform Baton in video fidelity and audio aesthetics, they show comparable prompt-following capabilities, showing the superiority of Baton in handling scenarios demanding semantic reasoning.

\subsection{Ablation Study}

\begin{table}[t!]\small
\caption{Ablation study on VA-Planner on Sem100. 
Only Planned Tokens removes global text embeddings and only injects planned tokens into the DiT. 
\textit{w/} Prompt Enhancement replaces VA-Planner with Qwen3-refined prompts. \textit{w/} Frozen LLM feeds frozen Qwen3-8B hidden states into DiT without VA-Planner. Tower refers to the dual semantic alignment towers. 
}
\vspace{-0.10in}
\begin{center}
\renewcommand\arraystretch{1.1}
\scalebox{0.85}{
\begin{tabular}{l|cccccc}
\toprule
Model                  & AQ$\uparrow$            & PQ$\uparrow$            & CU$\uparrow$            & M-WER$\downarrow$         & DeSync$\downarrow$        & P-Acc$\uparrow$         \\ \midrule
\textit{w/o} VA-Planner (Ovi~\citep{low2025ovi})         & 0.44          & 6.88          & 6.45          & 0.62          & 0.98          & 0.51          \\
Only Planned Tokens          & 0.53          & 7.38          & 7.04          & 0.17          & 0.73          & 0.78          \\
\textit{w/} Prompt Enhancement (PE-Qwen3-235B-A22B~\citep{yang2025qwen3})  & 0.51          & 6.97          & 6.64          & 0.56          & 0.93          & 0.62          \\
\textit{w/} Frozen LLM (Qwen3-8B~\citep{yang2025qwen3})          & 0.52          & 7.27          & 6.68          & 0.48          & 0.86          & 0.67          \\
\textit{w/o} Learnable Query    & 0.52          & 7.25          & 6.88          & 0.27          & 0.74          & 0.79          \\
\textit{w/o} Tower              & 0.48          & 6.92          & 6.53          & 0.45          & 0.78          & 0.69          \\
\textit{w/o} RoPE in Tower      & 0.42          & 6.57          & 6.24          & 0.68          & 1.05          & 0.44          \\
\textit{w/} TA-Tok~\citep{han2025vision}+WavTokenizer & 0.47          & 7.16          & 6.85          & 0.32          & 0.81          & 0.68          \\
\textit{w/} DINOv3~\citep{simeoni2025dinov3}              & 0.50          & 7.36          & 7.02          & 0.29          & 0.74          & 0.77          \\
\textit{w/} Beats~\citep{chen2022beats}               & 0.48          & 7.28          & 6.77          & 0.36          & 0.78          & 0.73          \\
\textit{w/} Unified Tokens~\citep{liu2025javisgpt}      & 0.44          & 7.12          & 6.73          & 0.41          & 0.85          & 0.65          \\ \midrule
Ours                   & \textbf{0.54} & \textbf{7.53} & \textbf{7.16} & \textbf{0.14} & \textbf{0.68} & \textbf{0.82} \\ \bottomrule
\end{tabular}
}
\end{center}
\label{table:ablation_va_planner}
\vspace{-0.10in}
\end{table}

\textbf{VA-Planner.}
We conduct an ablation study to validate the contributions of VA-Planner in Baton, as shown in Fig.\ref{fig:ablation} (a) and Table \ref{table:ablation_va_planner}.
Notably, all quantitative ablation studies are on the Sem100 dataset, and all ablated models are trained on the same dataset under identical settings.
\textit{w/} TA-Tok+WavTokenizer replaces the continuous SigLip2 and WavTokenizer targets with their discrete counterparts (TA-Tok~\citep{han2025vision} and discrete WavTokenizer). \textit{w/} DINOv3 replaces SigLip2 with DINOv3~\citep{simeoni2025dinov3} as the video alignment target. \textit{w/} Beats replaces WavTokenizer with Beats~\citep{chen2022beats} as the audio alignment target. \textit{w/} Unified Tokens~\citep{liu2025javisgpt} replaces separate video and audio planning with a single unified token sequence that jointly describes both modalities. 
By analyzing the results, we can obtain the following observations:
(1) Removing VA-Planner significantly degrades performance, particularly in M-WER, DeSync, and P-Acc, indicating that VA-Planner can significantly improve video-audio synchronization and prompt following ability by providing dedicated planned tokens. In contrast, \textit{w/} Prompt Enhancement and \textit{w/} Frozen LLM still suffer from accurate and stable video-audio generation due to their coarse guidance.
Notably, Only Planned Tokens already achieves strong results, confirming they are the primary semantic driver. The gap to the full model shows that global text embeddings and planned tokens are complementary. Global embeddings provide a holistic prior for coherence, while planned tokens supply fine-grained, position-specific structure. 
(2) Removing RoPE in Tower causes the most severe degradation, because the towers without temporal positional alignment produce temporally misaligned planned tokens that mislead denoising. 
\textit{w/o} Tower confirms that the video plan without bidirectional cross-modal attention is entirely blind to audio context (due to causal masking in the MLLM), producing two independently derived plans that lack mutual consistency and inevitably diverge during denoising.
\textit{w/o} Learnable Query shows that query-based distillation can gain more semantic details.
(3) Replacing continuous alignment targets with discrete TA-Tok~\citep{han2025vision}+WavTokenizer leads to clear drops across most metrics, because discrete quantization inevitably loses the fine-grained perceptual nuances that continuous features preserve, reducing the semantic richness available to guide each denoising step. 
\textit{w/} Unified Tokens~\citep{liu2025javisgpt} yields even larger degradation. Since video and audio possess fundamentally different spatio-temporal structures, collapsing them into a single token sequence forces both modalities to share one bottleneck, blurring modality-specific semantics and undermining the fine-grained cross-modal coordination that separate planning naturally provides.
(4) Both DINOv3~\citep{simeoni2025dinov3} and Beats~\citep{chen2022beats} underperform SigLip2 and WavTokenizer. It indicates that SigLip2's text-aligned visual features offer stronger semantic grounding than DINOv3's self-supervised features. While Beats is trained for audio classification and encodes category-level abstractions,  WavTokenizer's reconstruction-oriented features preserve the fine-grained acoustic details that the diffusion model requires for audio synthesis.

We further ablate the injection of planned tokens, keyframe FPS, and the orders of $T_{v}^{tag}$ and $T_{a}^{tag}$, as shown in Appx. \ref{sec: more_ablation_study}. 
Three key findings emerge.
(1) Our injection outperforms other alternatives, because a coarse-to-fine hierarchy lets the text prior and fine-grained planned cues complement rather than dilute each other.
(2) Keyframe FPS exhibits semantic saturation at FPS=6, as the keyframe-level blueprints already sufficiently cover the temporal semantics for denoising.
(3) Placing $T_{v}^{tag}$ before $T_{a}^{tag}$ exploits the MLLM's causal attention so that audio plan conditions on the video plan, matching the natural audio-follows-video dependency.
Details are in Appx.~\ref{sec: more_ablation_study}.

\begin{table}[t!]\small
\caption{Ablation study on RS-RoPE and different backbones. Temporal RoPE~\citep{team2026mova} only scales the position indices in the temporal axis. 
Baton-Qwen3-X uses Qwen3-X to initialize the MLLM. $\dagger$ denotes that single-GPU (80GB) inference is infeasible and 2-GPU FSDP parallelism is required. 
}
\vspace{-0.10in}
\begin{center}
\renewcommand\arraystretch{1.1}
\scalebox{0.75}{
\begin{tabular}{c|l|ccccccc}
\toprule
Category & Model                  & AQ$\uparrow$            & PQ$\uparrow$            & CU$\uparrow$            & M-WER$\downarrow$         & DeSync$\downarrow$        & P-Acc$\uparrow$         & Per-GPU Mem$\downarrow$        \\ \midrule
\multirow{2}{*}{RS-RoPE} & \textit{w/o} RS-RoPE            & 0.40          & 6.42          & 6.15          & 0.65          & 1.08          & 0.46          & 68.4G          \\ 
& \textit{w/} Temporal RoPE~\citep{team2026mova}       & 0.51          & 7.39          & 6.92          & 0.41          & 0.75          & 0.73          & 68.4G          \\ \midrule
\multirow{5}{*}{\makecell{MLLM\\Backbones}} & Baton-Qwen3-VL-8B-Instruct~\citep{Qwen3-VL}   & 0.44          & 6.90          & 6.51          & 0.45          & 0.84          & 0.72          & 71.2G          \\
& Baton-Qwen3-Omni-30B-A3B~\citep{xu2025qwen3_omni}     & 0.41          & 6.76          & 6.28          & 0.65          & 1.02          & 0.65          & 74.6G          \\
& Baton-Qwen3-4B~\citep{yang2025qwen3}               & 0.46          & 7.32          & 6.90          & 0.31          & 0.82          & 0.68          & 63.5G          \\
& Baton-Qwen3-30B-A3B-Instruct~\citep{yang2025qwen3} & 0.55          & 7.51          & 7.12          & 0.16          & 0.61          & 0.82 & 71.8GG          \\
& Baton-Qwen3-32B~\citep{yang2025qwen3}              & \textbf{0.58} & \textbf{7.86} & \textbf{7.29} & \textbf{0.13} & \textbf{0.54} & \textbf{0.85}          & 2$\times$69.4G$^\dagger$          \\ \midrule
\multirow{2}{*}{\makecell{DiT\\Backbones}} & LTX-2~\citep{hacohen2026ltx}                  & 0.48          & 7.14          & 6.91          & 0.58          & 0.97          & 0.62          & 48.5G          \\
& LTX-2~\citep{hacohen2026ltx}+Baton-Qwen3-8B (\textit{w/o} DiT)       & 0.56          & 7.49          & 7.21          & 0.16          & 0.64          & 0.83          & 64.6G          \\ \midrule
Final & Baton-Qwen3-8B~\citep{yang2025qwen3}                   & 0.54          & 7.53          & 7.16          & 0.14          & 0.68          & 0.82          & 68.4G          \\ \bottomrule
\end{tabular}
}
\end{center}
\label{table:ablation_rope_backbone}
\vspace{-0.10in}
\end{table}

\noindent\textbf{RS-RoPE.}
We conduct an ablation study on RS-RoPE, as shown in Table \ref{table:ablation_rope_backbone} and Fig. \ref{fig:ablation} (b). 
\textit{w/o} RS-RoPE causes drastic degradation, even worse than \textit{w/o} VA-Planner. The reason is that the cross-attention degenerates into purely content-based matching that treats all planned tokens as an orderless bag, collapsing the structured keyframe-level plan into a vague global signal that misleads denoising. 
\textit{w/} Temporal RoPE~\citep{team2026mova} recovers much of the performance but still falls short, as it only aligns the temporal axis while ignoring the spatial mismatch between the semantic grid and the latent grid, preventing latents from attending to their spatially corresponding planning cues.

\noindent\textbf{Different Backbones.}
We ablate the backbones of the VA-Planner and DiT, as shown in Fig. \ref{fig:ablation} (c) and Table \ref{table:ablation_rope_backbone}. 
We have the following observations:
(1) Qwen3-VL/Omni initialization are worse than ours, despite their larger pretraining scope. 
Qwen3-VL is pretrained for visual understanding with features heavily shaped toward high-level recognition rather than the spatial-perceptual structure required by SigLip2 alignment. 
Migrating such a distribution toward SigLip2's spatial domain is harder than training a general-purpose Qwen3 from scratch, as the pretrained weights resist the new alignment objective. 
Qwen3-Omni suffers similarly because its audio capabilities are pretrained on speech-centric data with a fixed voice identity, yielding an output distribution that lacks acoustic diversity. 
Forcing this narrow distribution to cover the broad range of environmental sounds and diverse timbres demands more data to overcome the pretrained bias.
(2) Scaling the MLLM backbone from 4B to 32B improves all metrics, confirming that stronger reasoning capacity leads to higher-quality planned tokens. 
However, Qwen3-32B incurs significant training and inference overhead. 
Qwen3-30B-A3B performs similarly to the dense Qwen3-8B despite more parameters, as MoE’s sparse routing activates different experts per token, fragmenting features and weakening the holistic cross-token reasoning needed for globally consistent long-horizon planning. Ours strikes a practical balance between performance and computational cost.
(3) Replacing the DiT backbone from Ovi~\citep{low2025ovi} to LTX-2~\citep{hacohen2026ltx} validates the robustness of our model across different DiT backbones.

\noindent\textbf{Training Stages.}
We compare our model in different training stages, as shown in Table \ref{table:ablation_training_stages}. We can see that skipping Stage-2 causes significant degradation. The root reason is that the DiT's cross-attention layers have never been exposed to semantic features. Directly receiving imperfect planner predictions in Stage-3 introduces a distribution shock that the model struggles to absorb. Our three-stage curriculum avoids this by letting Stage-2 first teach the DiT to leverage clean ground-truth features, establishing a stable semantic interface, then Stage-3 progressively adapts it to VA-Planner predictions, effectively reducing exposure bias.
Furthermore, Stage-2 outperforms our full model because it conditions on ground-truth SigLip2 and WavTokenizer features, which provide perfect semantic blueprints free from any prediction noise. However, ground-truth perceptual features are unavailable at inference since they require access to the target video and audio. Stage-3 closes this gap by adapting the DiT to realistic VA-Planner predictions, achieving robust generation under practical conditions.
More ablation studies are in Appx. \ref{sec: more_ablation_study}.

\subsection{Applications and User Study}

\textbf{Speed and GPU Resource.}
We compare the inference latency and GPU cost between Baton and previous models, as shown in Appx.\ref{sec: speed_gpu_resource_comparison}.
Compared to our DiT backbone Ovi~\citep{low2025ovi}, Baton introduces 28\% additional GPU memory, while achieving over 36\% improvement in DeSync and 78\% in P-Acc. It indicates that the planning overhead is a highly favorable trade-off for the substantial quality gains, highlighting its superiority in handling semantically complex scenarios.

\noindent\textbf{Multi-Speakers.}
We experiment on multi-speaker scenarios, as shown in Appx.\ref{sec: multiple_speaker_results}. We can see that Baton is capable of joint video-audio contents involving multi-character interaction.

\noindent\textbf{Complex Scene.}
We experiment in semantically complex scenarios, as shown in Appx. \ref{sec: complex_scene_results}. Baton can handle long-horizon scenes involving multiple sequential actions and complex interactions.


\noindent\textbf{User Study.}
We conduct a user study on 30 selected videos. The participants are university students and faculty. In each case, participants are first presented with the video/audio prompt. We then provide two videos, one of which is generated by Baton and the other is synthesized by a competitor.
Participants are asked to answer questions: V-A/A-A: "Which one has better video/audio alignment with the prompts?" I-C/B-C/V-A-S: "Which one has better ID/background consistency/video-audio synchronization?" 
Appx.\ref{sec: user_study_details} shows the superiority of our model in subjective evaluation.
\section{Conclusion}

In this paper, we proposed Baton, a video diffusion transformer with a dedicated MLLM to jointly synthesize stable video-audio, even for scenarios requiring complex semantic reasoning.
To explicitly disentangle semantic reasoning from synthesis before denoising, Baton first introduced the VA-Planner, an MLLM equipped with dual semantic alignment towers, which reasoned over user prompts to produce a pair of mutually aligned video and audio planned tokens as keyframe-level blueprints, anchoring both denoising trajectories to a shared semantic roadmap.
To address the spatio-temporal mismatch between planned tokens and diffusion latents, Baton further introduced Relative Semantic RoPE.
Extensive experiments demonstrated that Baton significantly outperforms open-source methods, particularly on complex prompts demanding semantic reasoning, and achieves comparable prompt-following capability to commercial models.

\bibliography{iclr2025_conference}
\bibliographystyle{iclr2025_conference}
\newpage
\clearpage

\appendix

\section{Appendix}

\subsection{Preliminaries}
\label{sec: preliminaries}

The diffusion model learns to generate data by reversing a noise corruption process. Following Rectified Flow~\citep{lipman2022flow}, the forward process linearly interpolates between a clean data sample $\bm{z}_{0}\sim\bm{p}_{\text{data}}$ and Gaussian noise $\bm{z}_{1}\sim\mathcal{N}(0,I)$:
\begin{equation}\small
\label{eq:forward_diffusion}
\begin{aligned}
    \bm{z}_{t}=(1-t)\bm{z}_{0}+t\bm{z}_{1},
\end{aligned}
\end{equation}
where $t \in [0,1]$ is a continuous timestep controlling the noise level. A neural network $\hat{\bm{v}}_{\theta}(\bm{z}_{t},t)$ is trained to regress the velocity field $\bm{z}_{1}-\bm{z}_{0}$ from the noisy sample $\bm{z}_{t}$ and timestep $t$, enabling iterative denoising from pure noise back to clean data at inference. The training objective is:
\begin{equation}\small
\label{eq:mse_loss}
\begin{aligned}
     \mathcal{L} = \mathbb{E}_{\bm{z}_{0},\bm{z}_{1},t}(\left \| (\bm{z}_{1}-\bm{z}_{0}) - \hat{\bm{v}}_{\theta}(\bm{z}_{t}, t)  \right \|^{2}).
\end{aligned}
\end{equation}

\subsection{Evaluation Metrics}
\label{sec: evaluation_metrics}

To comprehensively evaluate the model performance, we utilize multiple metrics to validate the model on Verse-Bench and Sem100 in terms of video quality, audio quality, video-audio synchronization, and prompt following accuracy.

\noindent\textbf{Video Quality.} We leverage imaging quality (IQ) and dynamic degree (DD) from VBench~\citep{huang2024vbench} to validate the synthesized video quality. 
We use the pretrained aesthetic-predictor-v2-5~\citep{aesthetic_predictor_v2_5} to predict the aesthetic quality (AQ)~\citep{hu2025harmony}.
We further evaluate the identity consistency (ID) by comparing the mean DINOv3~\citep{simeoni2025dinov3} representation between the first synthesized frame and the rest of the synthesized frames.

\noindent\textbf{Audio Quality.} We utilize the pretrained AudioBox~\citep{tjandra2025meta_audio_aesthetics} to evaluate the perceptual audio quality across production quality (PQ) and content usefulness (CU).
To assess the accuracy of multi-speaker speech, we propose the multi-speaker word error rate (M-WER). Given the synthesized audio, we first use Qwen3-Omni~\citep{xu2025qwen3_omni} to segment the speech into multiple transcripts by speaker identity based on vocal characteristics, and then compute the word error rate~\citep{radford2023robust} against the ground-truth transcript.

\noindent\textbf{Video-Audio Synchronization.}
We utilize the Sync-C~\citep{chung2016out} and Sync-D~\citep{chung2016out} to measure the synchronization of lips with audio. We further use the pretrained Synchformer~\citep{iashin2024synchformer} to quantify the temporal misalignment between video and audio streams.

\begin{figure}[t!]
\begin{center}
\includegraphics[width=1\linewidth]{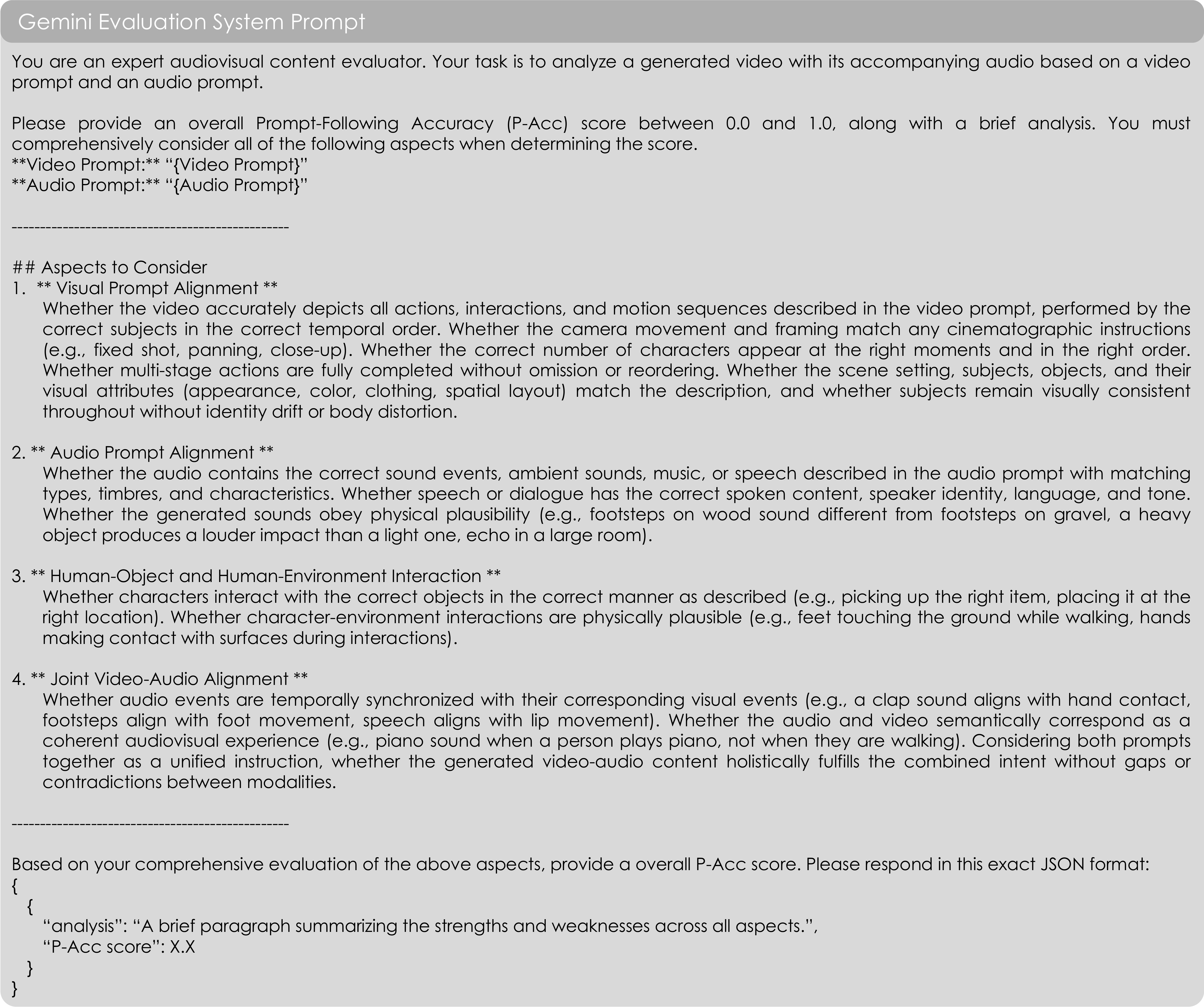}
\end{center}
\vspace{-0.3cm}
   \caption{The system prompt used for Gemini-based prompt following accuracy evaluation.
   }
\label{fig:p_acc_format}
\vspace{-0.25cm}
\end{figure}

\noindent\textbf{Prompt Following Accuracy.}
We utilize Gemini-3.1~\citep{gemini_3} to evaluate the prompt following accuracy of joint video-audio generation models. The evaluation prompts are shown in Fig. \ref{fig:p_acc_format}.

\subsection{Details of Dual Semantic Alignment Towers}
\label{sec: dual_semantic_alignment_tower_details}

In the dual semantic alignment towers, $\mathtt{CMAttn}_{v \rightarrow a}(\cdot)$ and $\mathtt{CMAttn}_{a \rightarrow v}(\cdot)$ require each query token from one modality to attend to temporally corresponding key tokens from the other modality. However, video and audio tokens have fundamentally different temporal densities. 
A video contains $N$ keyframes sampled at $\text{FPS}=6$, with each keyframe encoded into $n_v$ spatial tokens, yielding $L_v = N \times n_v$ tokens in total. 
Audio is represented as $M$ one-second chunks, with each chunk encoded into $n_a$ tokens, yielding $L_a = M \times n_a$ tokens. Without positional encoding, the cross-modal attention is content-based and cannot distinguish temporally aligned tokens from misaligned ones.

To address this, we assign a continuous timestamp $\tau$ to each token based on its actual temporal position, and apply Rotary Position Embedding (RoPE) using these timestamps in the $\mathtt{CMAttn}_{v \rightarrow a}(\cdot)$ and $\mathtt{CMAttn}_{a \rightarrow v}(\cdot)$. This maps both modalities onto a shared continuous time axis, so that the attention scores naturally reflect temporal proximity.

\noindent\textbf{Timestamp Assignment.}
Since the total video duration is $M$ seconds (equal to $M$ audio chunks), we assign each token a physical timestamp in seconds based on its position within the total duration.
For video, the $N$ keyframes are uniformly distributed across $M$ seconds, so the $i$-th keyframe ($i = 0, \dots, N-1$) corresponds to time $\tau_{v}^{(i, \cdot)} = i \cdot M / N$ seconds. Since the $n_v$ spatial tokens within a single frame represent different spatial patches of the same temporal instant, they share the same timestamp. Thus, for the $j$-th spatial token of the $i$-th keyframe:
\begin{equation}\small
\label{eq:video_timestamp_assignment}
\begin{aligned}
     \tau_v^{(i,j)} = \frac{i \cdot M}{N}, \quad i = 0, \dots, N\!-\!1, \;\; j = 0, \dots, n_v\!-\!1,
\end{aligned}
\end{equation}
For audio, the $m$-th chunk ($m = 0, \dots, M-1$) covers the time interval $[m, m\!+\!1)$ seconds. Unlike video spatial tokens that share the same temporal instant, the $n_a$ tokens within a chunk are arranged in temporal order (each corresponding to a successive temporal position in the encoder output), so they receive evenly spaced timestamps within the chunk:
\begin{equation}\small
\label{eq:audio_timestamp_assignment}
\begin{aligned}
     \tau_a^{(m,j)} = m + \frac{j}{n_a}, \quad m = 0, \dots, M\!-\!1, \;\; j = 0, \dots, n_a\!-\!1,
\end{aligned}
\end{equation}

\noindent\textbf{RoPE Application.}
Given token embeddings $\bm{h} \in \mathbb{R}^{D}$ and its timestamp $\tau$, we apply RoPE as:
\begin{equation}\small
\label{eq:rope_application_tower}
\begin{aligned}
     \mathtt{RoPE}(\bm{h}, \tau) = \bm{h} \odot \cos(\tau \cdot \boldsymbol{\omega}) + \mathtt{rotate}(\bm{h}) \odot \sin(\tau \cdot \boldsymbol{\omega}),
\end{aligned}
\end{equation}
where $\boldsymbol{\omega} \in \mathbb{R}^{D/2}$ is the frequency vector with $\omega_k = \frac{1}{\theta^{2k/D}}$ ($\theta$ is a base frequency hyperparameter).
$\odot$ denotes element-wise multiplication and $\mathtt{rotate}(\cdot)$ swaps adjacent pairs with sign flips following the standard RoPE.

The dual towers involve two symmetric cross-modal attention operations. Let $\hat{\bm{F}}_v = \mathtt{CAttn}_v(\bm{Q}_v, \bm{H}_v)$ and $\hat{\bm{F}}_a = \mathtt{CAttn}_a(\bm{Q}_a, \bm{H}_a)$ denote the intermediate features from the intra-modal cross-attention of each tower. Both CAttn operations execute first, and then the two towers exchange features via CMAttn. In the video tower $\mathtt{CMAttn}_{a \rightarrow v}(\cdot)$, $\hat{\bm{F}}_v$ serves as queries and $\hat{\bm{F}}_a$ as keys/values. In the audio tower $\mathtt{CMAttn}_{v \rightarrow a}(\cdot)$, the roles are reversed. Both directions share the same timestamp assignment, with RoPE applied to queries and keys using their respective timestamps:
\begin{equation}
\begin{aligned}
\mathtt{CMAttn}_{a \rightarrow v}(\hat{\bm{F}}_v, \hat{\bm{F}}_a) &= \mathtt{Attn}\big(\mathtt{RoPE}(\hat{\bm{F}}_v,\, \tau_v),\;\mathtt{RoPE}(\hat{\bm{F}}_a,\, \tau_a),\;\hat{\bm{F}}_a\big), \\
\mathtt{CMAttn}_{v \rightarrow a}(\hat{\bm{F}}_a, \hat{\bm{F}}_v) &= \mathtt{Attn}\big(\mathtt{RoPE}(\hat{\bm{F}}_a,\, \tau_a),\;\mathtt{RoPE}(\hat{\bm{F}}_v,\, \tau_v),\;\hat{\bm{F}}_v\big).
\end{aligned}
\end{equation}
Notably, since audio has no spatial structure, spatial positional encoding is not applied in CMAttn. All $n_v$ spatial tokens within the same video frame share the same timestamp, allowing them to attend uniformly to the same set of temporally corresponding audio tokens.

\subsection{Dataset Details}
\label{sec: dataset_details}

Regarding the training dataset, our training dataset (1.5 million video-audio clips, FPS=24) is aggregated from OpenHuman-Vid~\citep{li2025openhumanvid}, AudioCaps~\citep{kim2019audiocaps}, WavCaps~\citep{mei2024wavcaps}, and videos collected from the internet.
We apply a multi-stage filtering pipeline to ensure data quality. First, we use aesthetic-predictor-v2-5~\citep{aesthetic_predictor_v2_5} to predict the visual aesthetic quality of each video and retain only clips with a score above 0.4. We then filter out static or near-static videos by requiring a Dynamic Degree~\citep{huang2024vbench} above 0.2. For audio quality, we use AudioBox~\citep{tjandra2025meta_audio_aesthetics} to evaluate audio aesthetics and keep only clips with PQ above 6.0. Finally, for videos containing speech, we apply SyncNet~\citep{chung2016out} to assess lip-sync accuracy and retain only clips with a confidence score above 0.9.
The resulting dataset covers a diverse range of audiovisual scenarios, including single-speaker and multi-speaker speech or singing, natural environmental sounds (e.g., wind, rain, animal calls), sci-fi and virtual environment sounds, and human-object or human-environment interaction scenes (e.g., cooking and instrument playing). 
Furthermore, we utilize 
Qwen3-VL-235B-A22B-Instruct~\citep{Qwen3-VL} and Qwen3-Omni~\citep{xu2025qwen3_omni} to caption the training video dataset to gain the video prompts and audio prompts, respectively.

\begin{figure}[t!]
\begin{center}
\includegraphics[width=1\linewidth]{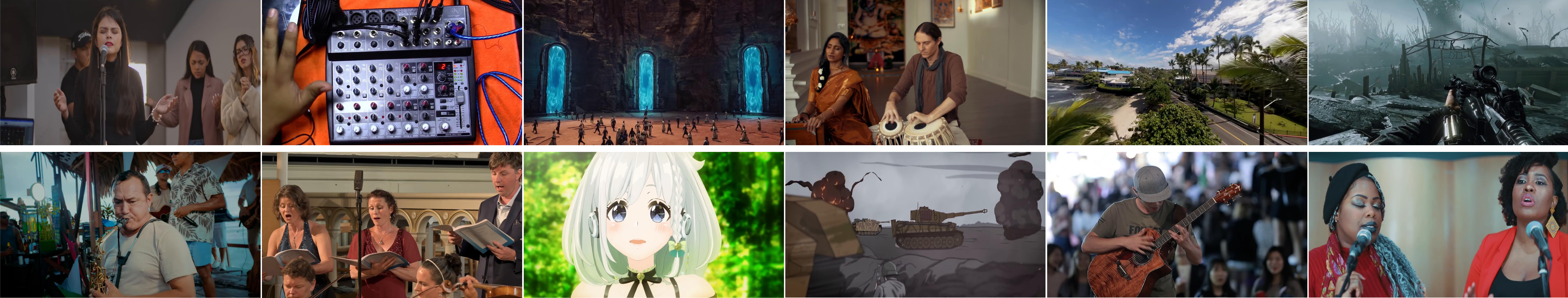}
\end{center}
\vspace{-0.3cm}
   \caption{Examples from Sem100.
   }
\label{fig:sem100}
\vspace{-0.25cm}
\end{figure}

In terms of the testing dataset, we select 100 unseen videos (10 seconds long, FPS=24) from the internet to construct the testing dataset Sem100. 
Some ground truth examples are shown in Fig. \ref{fig:sem100}.
The sources of videos come from numerous social media platforms, including YouTube and BiliBili. These videos feature individuals across diverse ethnicities, genders, and age groups, portrayed in full-body, half-body, and close-up shots against varied indoor and outdoor settings. Sem100 covers a broad spectrum of audiovisual scenarios, including single-speaker and multi-speaker dialogues, singing performances, natural environmental sounds, human-object interactions (e.g., cooking, sports, musical instruments), and human-environment interactions in both realistic and stylized scenes. 
In contrast to existing open-source testing datasets (Verse-Bench), the text prompts of Sem100 are deliberately designed to demand complex semantic reasoning, involving multi-stage sequential actions, compositional instructions, and intricate causal relationships across both visual and auditory modalities.
The full text prompts are saved in the JSON file in the supplementary material (zip).
The licenses of Verse-Bench and Sem100 are both Apache-2.0.

\subsection{Training Details}
\label{sec: training_details}

\begin{figure}[t!]
\begin{center}
\includegraphics[width=1\linewidth]{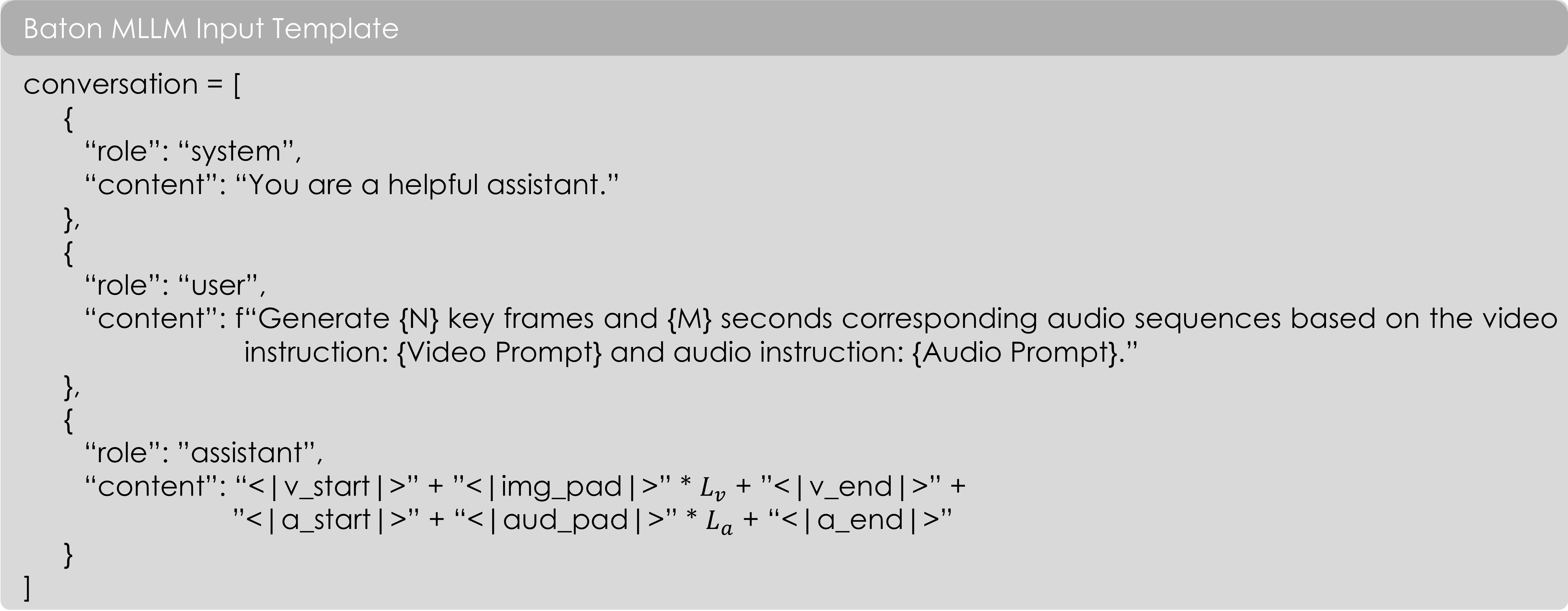}
\end{center}
\vspace{-0.3cm}
   \caption{The MLLM input template (system prompt, video prompt, and audio prompt).
   }
\label{fig:mllm_template}
\vspace{-0.25cm}
\end{figure}

We use the AdamW optimizer with parameters $\beta_{1}=0.9$, $\beta_{2}=0.999$. The model is trained at bf16 precision, equipped with DeepSpeed-Stage-3 for distributed data parallel training.
The training process contains three stages, and the learning rates in all stages are $1e-5$.
The learnable queries $\bm{Q}_v \in \mathbb{R}^{L_v \times D}$ and $\bm{Q}_a \in \mathbb{R}^{L_a \times D}$ in the dual semantic alignment towers are initialized from a normal distribution ($\sigma\!=\!0.02$) and have the same sequence length as the planned video and audio tokens ($L_v$ and $L_a$), respectively.
The training video dataset in the 3 training stages is the same.
Fig. \ref{fig:mllm_template} shows the MLLM input template during training.
In particular, in the first training stage (VA-Planner Pretraining), the MLLM is initialized by Qwen3-8B~\citep{yang2025qwen3}, and the whole MLLM and $\mathtt{SMLP}_{v/a}(\cdot)$ are trainable.
The frozen video semantic encoder is initialized by SigLip2~\citep{tschannen2025siglip} (siglip2-so400m-patch14-384), and the frozen audio semantic encoder is initialized by WavTokenizer~\citep{ji2024wavtokenizer} (WavTokenizer-large-unify-40token). We train the MLLM for 10 epochs, with a batch size of 1 per GPU.
In the second training stage (DiT Semantic Adaptation), the DiT is initialized by Ovi~\citep{low2025ovi}. DiT and $\mathtt{LMLP}_{v/a}(\cdot)$ remain trainable. We train the DiT for 10 epochs in 640p videos (FPS=24), with a batch size of 1 per GPU.
In the third training stage, only DiT remains trainable, and MLLM is frozen. We still train the DiT for 10 epochs in 640p videos (FPS=24), with a batch size of 1 per GPU.

\subsection{More Comparison Results}
\label{sec: more_comparison_results}

\begin{figure}[t!]
\begin{center}
\includegraphics[width=0.95\linewidth]{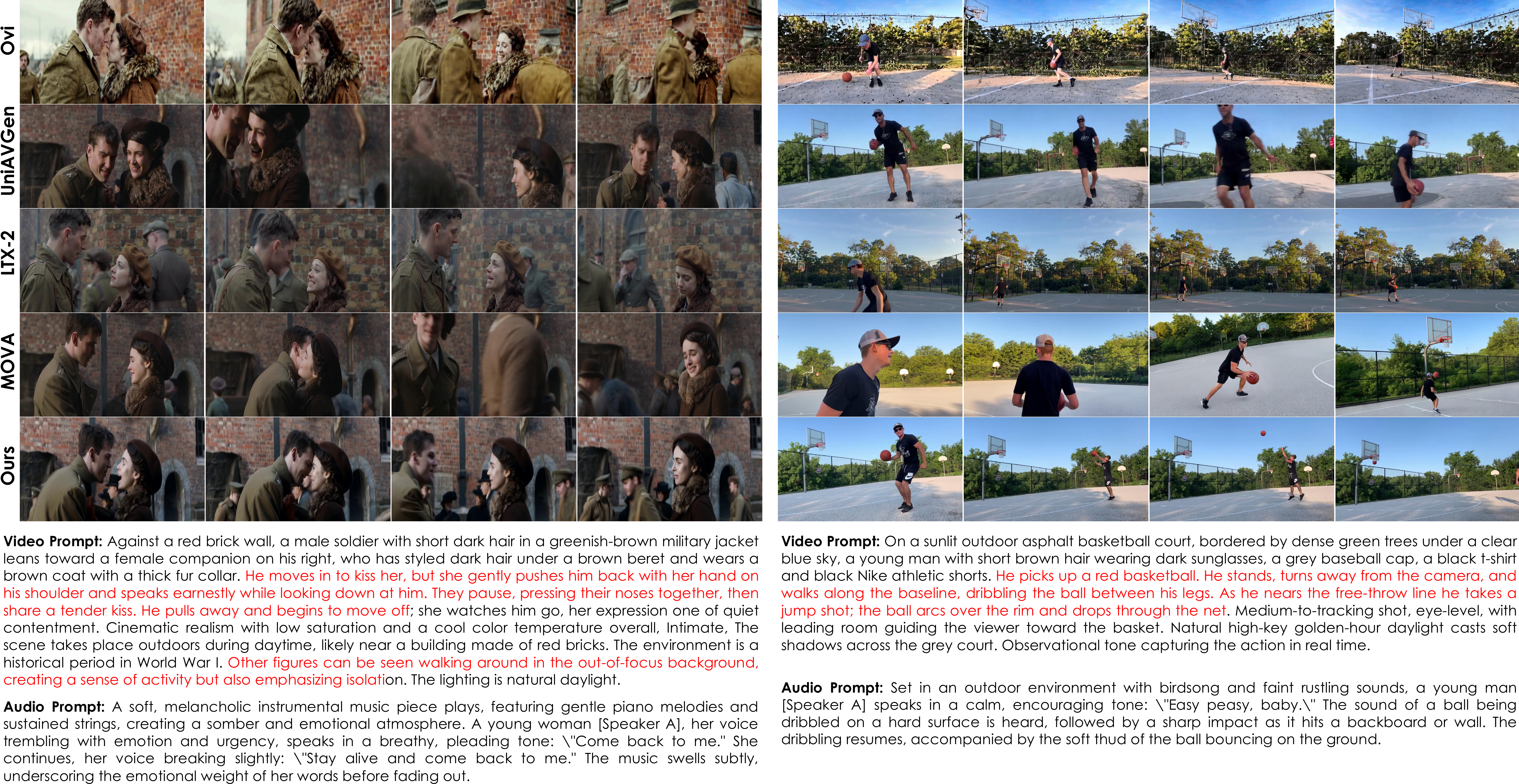}
\end{center}
\vspace{-0.3cm}
   \caption{More comparison results (1/3). Please refer to the demo video for audio.
   }
\label{fig:comparison_supp_1}
\vspace{-0.35cm}
\end{figure}

\begin{figure}[t!]
\begin{center}

\includegraphics[width=0.95\linewidth]{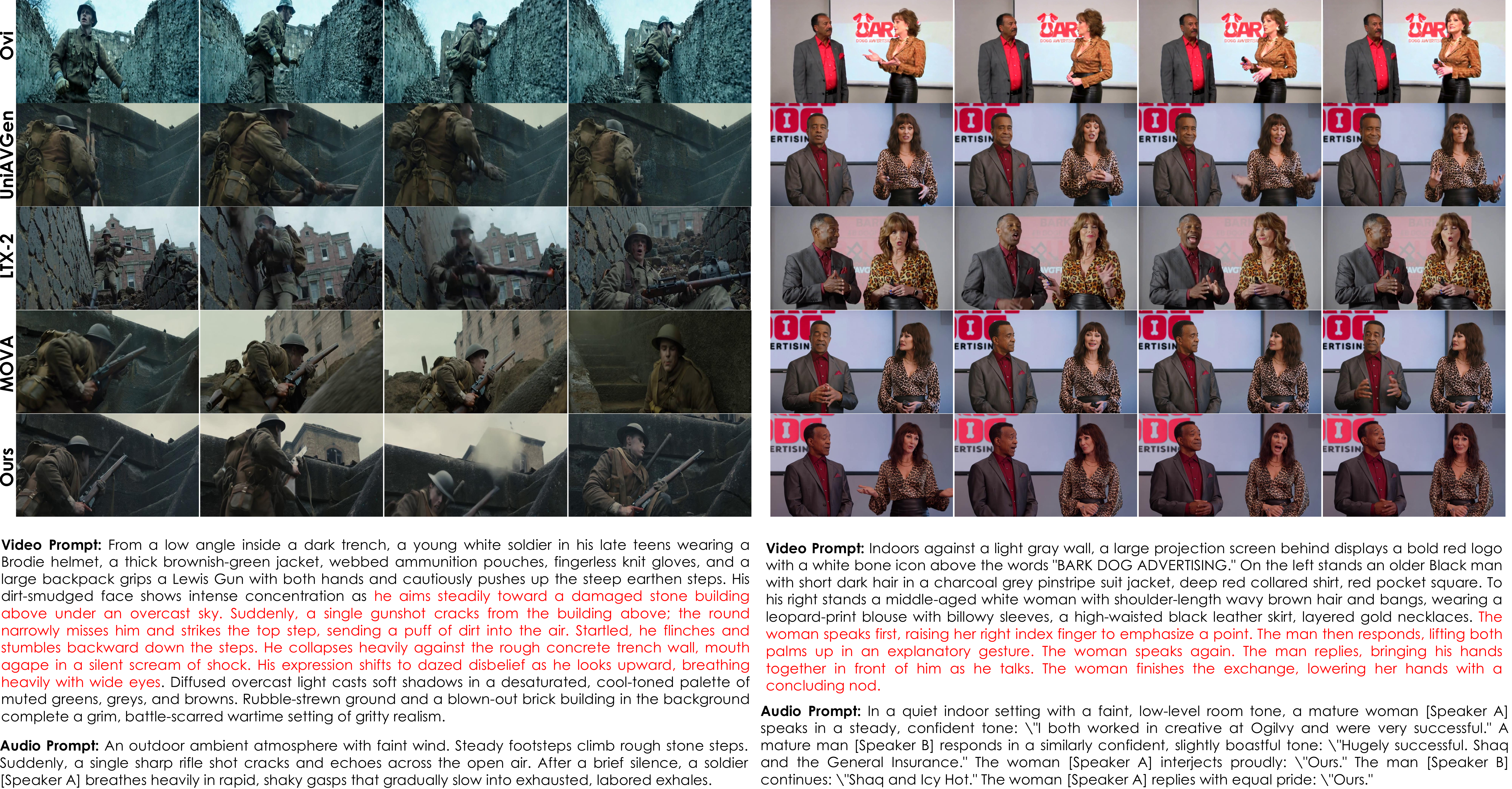}
\end{center}
\vspace{-0.3cm}
   \caption{More comparison results (2/3). Please refer to the demo video for audio.
   }
\label{fig:comparison_supp_2}
\vspace{-0.35cm}
\end{figure}

\begin{figure}[t!]
\begin{center}
\includegraphics[width=0.95\linewidth]{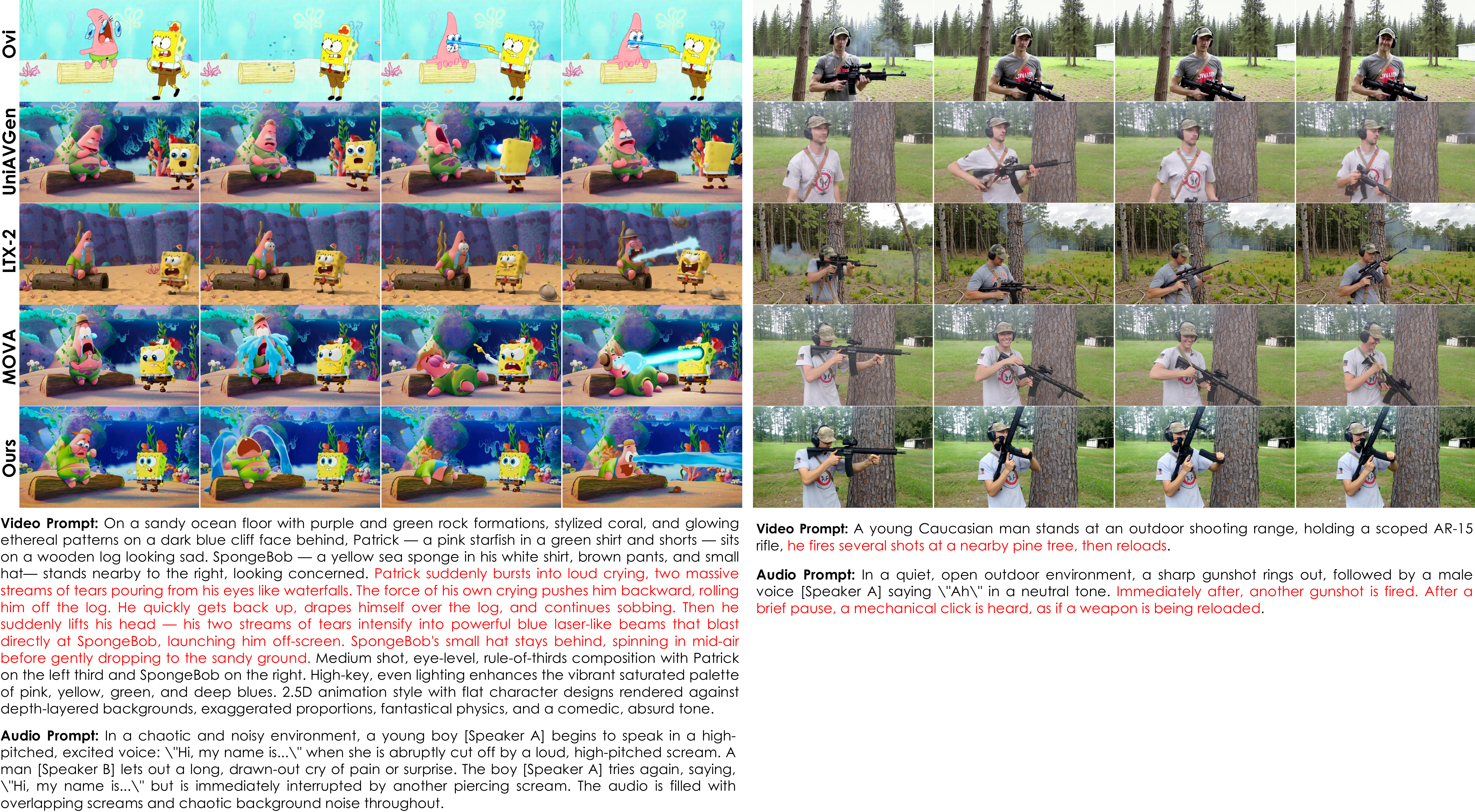}
\end{center}
\vspace{-0.3cm}
   \caption{More comparison results (3/3). Please refer to the demo video for audio.
   }
\label{fig:comparison_supp_3}
\vspace{-0.35cm}
\end{figure}

Fig. \ref{fig:comparison_supp_1}, Fig. \ref{fig:comparison_supp_2}, and Fig. \ref{fig:comparison_supp_3} show additional comparison results between our Baton and the state-of-the-art open-source models.
While previous models~\citep{low2025ovi, zhang2025uniavgen, hacohen2026ltx, team2026mova} often struggle with user prompts involving complex sequences of actions, multi-person dialogue, and intricate interactions between characters and their environment, they tend to misunderstand or omit key narrative details. As a result, the generated audio and video may not align with the prompt or may even be unsynchronized with each other. In some cases, discrepancies in how the audio and video branches interpret coarse global text embeddings can lead to conflicts during joint generation, causing visual artifacts such as body distortion or audio issues like noise and unnatural effects. In contrast, Baton accurately follows user prompts and produces stable, high-quality, and well-synchronized video-audio outputs.

\begin{table}[t!]\small
\caption{Quantitative comparison results between Baton and commercial models.
}
\vspace{-0.10in}
\begin{center}
\renewcommand\arraystretch{1.1}
\scalebox{0.85}{
\begin{tabular}{lccccccc}
\toprule
Model       & AQ$\uparrow$            & DD$\uparrow$            & PQ$\uparrow$            & CU$\uparrow$            & M-WER$\downarrow$         & DeSync$\downarrow$        & P-Acc$\uparrow$         \\ \midrule
Veo3.1~\citep{veo3_2025}      & 0.58          & 0.53          & 8.27          & 7.96          & 0.14          & 0.35          & 0.75          \\
Wan2.7~\citep{wan2_7}      & 0.63          & 0.62          & 8.18          & 7.75          & 0.19          & 0.38          & 0.85          \\
Kling3.0~\citep{kling3}    & 0.64          & 0.56          & 8.12          & 7.82          & 0.13          & 0.45          & 0.72          \\
Seedance2.0~\citep{seedance2_2025} & \textbf{0.68} & \textbf{0.68} & \textbf{8.54} & \textbf{8.07} & \textbf{0.11} & \textbf{0.28} & \textbf{0.92} \\ \midrule
Ours        & 0.54          & 0.48          & 7.53          & 7.16          & 0.14          & 0.68          & 0.82          \\ \bottomrule
\end{tabular}
}
\end{center}
\label{table:commercial_comparison}
\vspace{-0.10in}
\end{table}

\subsection{Comparison with Commercial Models}
\label{sec: comparison_commercial_models}

\begin{figure}[t!]
\begin{center}
\includegraphics[width=0.95\linewidth]{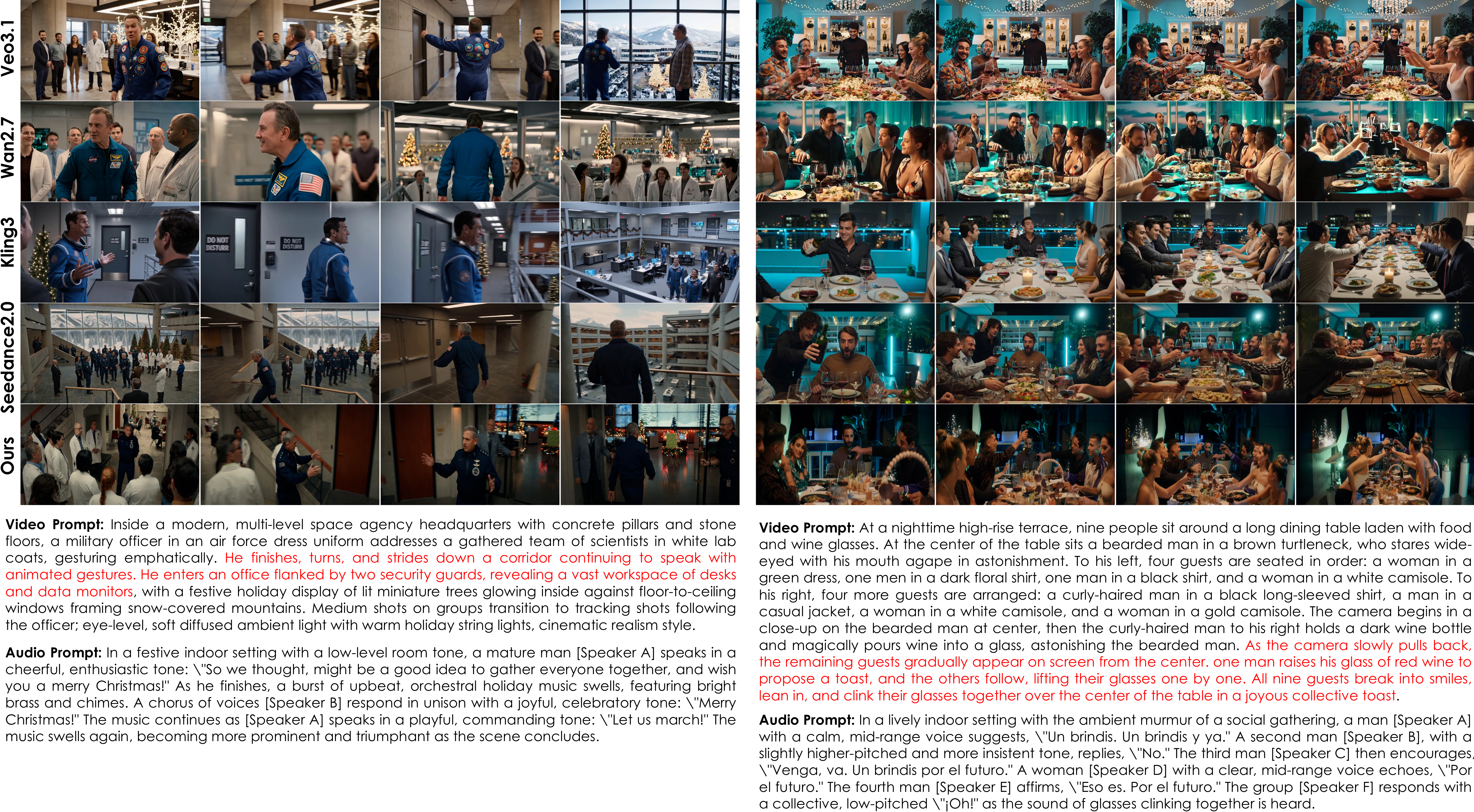}
\end{center}
\vspace{-0.3cm}
   \caption{Comparison results between Baton and commercial models. Please refer to the demo video for audio.
   }
\label{fig:commercial_comparison}
\vspace{-0.25cm}
\end{figure}

Table \ref{table:commercial_comparison} and Fig. \ref{fig:commercial_comparison} depict the comparison results between our Baton and closed-source commercial models.
Although Baton lags behind commercial models in visual quality and audio aesthetics, it achieves comparable performance in prompt following. For instance, in Fig. \ref{fig:commercial_comparison}, the right-side case shows that Veo 3.1 generates an incorrect number of people, Kling 3.0 and Wan 2.7 fail to match the specified positions of the pouring character, and Seedance 2.0 produces an incorrect order of character appearances. By contrast, Baton faithfully generates video-audio content that aligns with the text prompt, demonstrating its superiority in handling complex prompts that demand semantic reasoning.

\subsection{More Ablation Study}
\label{sec: more_ablation_study}

\textbf{Ablation on Planned Token Injection.}
We additionally conduct an ablation study on the injection of planned tokens predicted by our VA-Planner. The results are depicted in Table \ref{table:ablation_planned_token_injection}. We can see that our injection achieves the best performance. Concatenating planned tokens with text embeddings forces both signals to share the same attention layer, where the localized spatio-temporal semantics of the planned tokens are diluted by the coarse global text signal, degrading DeSync and P-Acc. In parallel injection, both signals independently modify the raw latents without hierarchical interaction, losing the chance for semantic complementarity. Our cascaded design first lets text cross-attention establish a coarse semantic prior, then V/ACross-Attn refines it with fine-grained, position-specific planning cues, yielding the best performance.

\begin{table}[t!]\small
\caption{Ablation study on the injection of planned tokens. Concat with Text Embeddings refers to concatenating the planned tokens with the global text embeddings. Parallel TCAttn+V/ACAttn performs TCross-Attn and V/ACross-Attn in parallel, rather than the original cascaded design.
}
\vspace{-0.10in}
\begin{center}
\renewcommand\arraystretch{1.1}
\scalebox{0.85}{
\begin{tabular}{l|cccccc}
\toprule
Model                  & AQ$\uparrow$            & PQ$\uparrow$            & CU$\uparrow$            & M-WER$\downarrow$         & DeSync$\downarrow$        & P-Acc$\uparrow$         \\ \midrule
Concat with Text Embeddings & 0.50          & 7.45          & 7.01          & 0.17          & 0.77          & 0.76          \\
Parallel TCAttn+V/ACAttn    & 0.52          & 7.48          & 7.14          & 0.14          & 0.71          & 0.79          \\ \midrule
Ours                        & \textbf{0.54} & \textbf{7.53} & \textbf{7.16} & \textbf{0.14} & \textbf{0.68} & \textbf{0.82} \\ \bottomrule
\end{tabular}
}
\end{center}
\label{table:ablation_planned_token_injection}
\vspace{-0.10in}
\end{table}

\begin{table}[t!]\small
\caption{Ablation study on keyframe FPS in the planning stage.
}
\vspace{-0.10in}
\begin{center}
\renewcommand\arraystretch{1.1}
\scalebox{0.85}{
\begin{tabular}{lcccccccc}
\toprule
Model        & AQ$\uparrow$            & DD$\uparrow$            & PQ$\uparrow$            & CU$\uparrow$            & M-WER$\downarrow$         & DeSync$\downarrow$        & P-Acc$\uparrow$         & GPU Mem$\downarrow$        \\ \midrule
FPS=1        & 0.44          & 0.36          & 7.18          & 6.81          & 0.46          & 0.78          & 0.70          & 59.6G          \\
FPS=4        & 0.47          & 0.41          & 7.35          & 6.92          & 0.33          & 0.72          & 0.78          & 62.7G          \\
FPS=6 (Ours) & 0.54          & 0.48          & 7.53          & 7.16          & 0.14          & 0.68          & 0.82          & 68.4G          \\
FPS=8        & \textbf{0.57} & \textbf{0.49} & \textbf{7.81} & \textbf{7.23} & \textbf{0.12} & \textbf{0.59} & \textbf{0.85} & \textbf{74.2G} \\
FPS=12       & -             & -             & -             & -             & -             & -             & -             & OOM            \\ \bottomrule
\end{tabular}
}
\end{center}
\label{table:ablation_keyframe_fps}
\vspace{-0.10in}
\end{table}

\begin{table}[t!]\small
\caption{Ablation on Orders of $T_{v}^{tag}$ and $T_{a}^{tag}$.
}
\vspace{-0.10in}
\begin{center}
\renewcommand\arraystretch{1.1}
\scalebox{0.85}{
\begin{tabular}{lcccccc}
\toprule
Model  & AQ$\uparrow$            & PQ$\uparrow$            & CU$\uparrow$            & M-WER$\downarrow$         & DeSync$\downarrow$        & P-Acc$\uparrow$         \\ \midrule
Interleave \texttt{<|img\_pad|>} and \texttt{<|aud\_pad|>}~\citep{xu2025qwen3_omni} & 0.47          & 7.24          & 6.83          & 0.31          & 0.75          & 0.75          \\
$[T^{tag}_{a}; T^{tag}_{v}]$                                                                                                               & 0.50          & 7.31          & 6.97          & 0.25          & 0.73          & 0.79          \\ \midrule
$[T^{tag}_{v}; T^{tag}_{a}]$ (Ours)                                                                                                        & \textbf{0.54} & \textbf{7.53} & \textbf{7.16} & \textbf{0.14} & \textbf{0.68} & \textbf{0.82} \\ \bottomrule
\end{tabular}
}
\end{center}
\label{table:ablation_order_placeholder}
\vspace{-0.10in}
\end{table}

\begin{table}[t!]\small
\caption{Ablation study on training stages. 
Stage-2 directly utilizes the groundtruth SigLip2 embeddings and WavTokenizer embeddings as the planned tokens, and injects them into DiT without MLLM prediction.
Skip Stage-2 refers to omitting Stage-2 training, and the training process includes only Stage-1 and Stage-3. 
}
\vspace{-0.10in}
\begin{center}
\renewcommand\arraystretch{1.1}
\scalebox{0.85}{
\begin{tabular}{l|cccccc}
\toprule
Model                  & AQ$\uparrow$            & PQ$\uparrow$            & CU$\uparrow$            & M-WER$\downarrow$         & DeSync$\downarrow$        & P-Acc$\uparrow$         \\ \midrule
Stage-2      & \textbf{0.60} & \textbf{7.92} & \textbf{7.70} & \textbf{0.12} & \textbf{0.45} & \textbf{0.87} \\
Skip Stage-2 & 0.49          & 7.26          & 6.73          & 0.40          & 0.82          & 0.75          \\ \midrule
Ours         & 0.54          & 7.53          & 7.16          & 0.14          & 0.68          & 0.82          \\ \bottomrule
\end{tabular}
}
\end{center}
\label{table:ablation_training_stages}
\vspace{-0.10in}
\end{table}

\begin{table}[t!]\small
\caption{Ablation study on planned token prediction accuracy.
}
\vspace{-0.10in}
\begin{center}
\renewcommand\arraystretch{1.1}
\scalebox{0.85}{
\begin{tabular}{lccccccc}
\toprule
Model                        & AQ$\uparrow$            & PQ$\uparrow$            & CU$\uparrow$            & M-WER$\downarrow$         & DeSync$\downarrow$        & P-Acc$\uparrow$         & MSE Loss$\downarrow$      \\ \midrule
w/o Learnable Query          & 0.52          & 7.25          & 6.88          & 0.27          & 0.74          & 0.79          & 0.37          \\
w/o Tower                    & 0.48          & 6.92          & 6.53          & 0.45          & 0.78          & 0.69          & 0.39          \\
w/o RoPE in Tower            & 0.42          & 6.57          & 6.24          & 0.68          & 1.05          & 0.44          & 0.46          \\
Baton-Qwen3-VL-8B-Instruct   & 0.44          & 6.90          & 6.51          & 0.45          & 0.84          & 0.72          & 0.45          \\
Baton-Qwen3-Omni-30B-A3B     & 0.41          & 6.76          & 6.28          & 0.65          & 1.02          & 0.65          & 0.50          \\
Baton-Qwen3-4B               & 0.46          & 7.32          & 6.90          & 0.31          & 0.82          & 0.68          & 0.41          \\
Baton-Qwen3-32B              & \textbf{0.58} & \textbf{7.86} & \textbf{7.29} & \textbf{0.13} & \textbf{0.54} & \textbf{0.85} & \textbf{0.27} \\
Baton-Qwen3-30B-A3B-Instruct & 0.55          & 7.51          & 7.12          & 0.16          & 0.61          & 0.82          & 0.35          \\ \midrule
Ours (Baton-Qwen3-8B)        & 0.54          & 7.53          & 7.16          & 0.14          & 0.68          & 0.82          & 0.31          \\ \bottomrule
\end{tabular}
}
\end{center}
\label{table:ablation_plan_prediciton}
\vspace{-0.10in}
\end{table}

\noindent\textbf{Ablation on Keyframe FPS in Planning Stage.}
We ablate the keyframe FPS in the planning stage, as shown in Table \ref{table:ablation_keyframe_fps}. 
We observe that as the keyframe FPS increases, more keyframe blueprints are provided to the DiT, leading to improved performance. 
However, GPU memory consumption also rises significantly. At FPS = 12, single-GPU inference exceeds 80GB and results in Out-Of-Memory (OOM) errors. Notably, the performance gain from FPS = 8 over FPS = 6 is smaller than that from FPS = 4 to FPS = 6. We attribute this to semantic saturation. At FPS = 6, the keyframe-level semantic cues provided to the DiT are already sufficient, and further increasing the number of keyframes introduces redundant information rather than additional useful semantics. To balance computational cost and model performance, we therefore choose FPS = 6 as the final configuration.

\noindent\textbf{Ablation on Orders of $T_{v}^{tag}$ and $T_{a}^{tag}$.}
We ablate the position orders of $T_{v}^{tag}$ and $T_{a}^{tag}$ in the MLLM input, as shown in Table \ref{table:ablation_order_placeholder}.
Interleave \texttt{<|img\_pad|>} and \texttt{<|aud\_pad|>}~\citep{xu2025qwen3_omni} follows the Qwen3-Omni~\citep{xu2025qwen3_omni}, which interleaves \texttt{<|img\_pad|>} and \texttt{<|aud\_pad|>} at the chunk level so that temporally co-occurring video and audio tokens are adjacent.
Despite providing explicit temporal proximity, interleaving fragments each modality's token sequence, preventing the MLLM from building coherent intra-modal representations before cross-modal reasoning, and yields the worst results.
$[T^{tag}_{a}; T^{tag}_{v}]$ places audio before video, which improves over interleaving but still underperforms Ours. 
In the causal attention of MLLM, audio tokens can only condition on the text prompt without any visual context, so the audio plan is generated blind to the visual scene. The subsequent video tokens do attend to the audio plan, but this inverts the natural dependency. In most scenarios, sounds accompany and react to visual events rather than the reverse.
Our $[T^{tag}_{v}; T^{tag}_{a}]$ ordering exploits this asymmetry by letting the MLLM first build a complete video plan from the text prompt, then conditioning audio planning on both the prompt and the full video plan via causal attention. This produces audio tokens that are inherently synchronized with the visual blueprint.

\noindent\textbf{Ablation on Training Stages.}
We conduct an ablation study on training stages, as shown in Table \ref{table:ablation_training_stages}. The results indicate that Stage 3 acts as a critical adaptation phase, aligning training conditions with inference-time planner outputs. By replacing ideal encoder features with realistic predictions, it reduces exposure bias and significantly improves robustness during generation.

\noindent\textbf{Ablation on Planned Token Prediction Accuracy.}
We conduct an ablation study on planned token prediction accuracy, as shown in Table \ref{table:ablation_plan_prediciton}. 
In particular, for the MSE Loss column, we take each Stage-1-trained MLLM variant and run it on the training set. We extract ground-truth semantic embeddings from frozen SigLip2 and WavTokenizer encoders and compute the MSE against the MLLM-predicted planned tokens, thereby measuring how accurately the VA-Planner learns to approximate the target perceptual features after Stage-1 pretraining. All other metrics are evaluated on Sem100 using the full three-stage model.
The results reveal a strong positive correlation between Stage-1 prediction accuracy (MSE) and final generation quality. 
Qwen3-32B achieves the lowest MSE and the best downstream metrics, confirming that stronger language model reasoning capacity produces more accurate planned tokens, which in turn provides higher-quality blueprints for the DiT. Conversely, \textit{w/o} RoPE in Tower and Qwen3-Omni exhibit the high MSE, and correspondingly suffer the most severe degradation in P-Acc and DeSync, indicating that temporally misaligned or distribution-biased planned tokens mislead rather than guide the denoising process.
Notably, the correlation is not strictly monotonic. Ours achieves comparable metrics to Baton-Qwen3-30B-A3B despite similar prediction accuracy. Since MSE only measures pointwise token-level regression error, it cannot capture the inter-token temporal coherence across the planned token sequence. As discussed in the backbone ablation, MoE's sparse routing activates different experts per token, fragmenting features and weakening holistic cross-token reasoning. Consequently, even when the average per-token MSE is similar, the planned tokens from MoE may lack the globally consistent temporal structure that a dense model naturally maintains through unified parameter sharing, and this structural coherence is critical for guiding long-horizon denoising. 
Furthermore, since the planned token prediction is not perfect, Stage-3 joint fine-tuning can effectively compensate for planner imperfections by adapting the DiT to realistic prediction noise. However, it cannot rescue severely inaccurate plans (e.g., \textit{w/o} RoPE in Tower), highlighting that a minimum level of Stage-1 prediction quality is necessary for the three-stage curriculum to succeed.

\begin{table}[t!]\small
\caption{Inference speed and GPU memory consumption comparison results.
}
\vspace{-0.10in}
\begin{center}
\renewcommand\arraystretch{1.1}
\scalebox{0.85}{
\begin{tabular}{l|cccccccc}
\toprule
Model                  & AQ$\uparrow$            & PQ$\uparrow$            & CU$\uparrow$            & M-WER$\downarrow$         & DeSync$\downarrow$        & P-Acc$\uparrow$         & GPU Mem$\downarrow$        & Speed$\downarrow$        \\ \midrule
Ovi~\citep{low2025ovi}      & 0.42          & 6.74          & 6.23          & 0.66          & 1.06          & 0.46          & 53.6G          & 184s         \\
UniAVGen~\citep{zhang2025uniavgen} & 0.31          & 6.68          & 5.83          & 0.72          & 1.14          & 0.44          & 46.8G          & 306s         \\
LTX-2~\citep{hacohen2026ltx}    & 0.48          & 7.14          & 6.91          & 0.58          & 0.97          & 0.62          & 48.5G          & 196s         \\
MOVA~\citep{team2026mova}     & 0.44          & 6.95          & 6.94          & 0.61          & 1.05          & 0.47          & 76.7G          & 1517s        \\ \midrule
Ours     & \textbf{0.54} & \textbf{7.53} & \textbf{7.16} & \textbf{0.14} & \textbf{0.68} & \textbf{0.82} & 68.4G          & 346s         \\ \bottomrule
\end{tabular}
}
\end{center}
\label{table:speed_gpu_resource_comparison}
\vspace{-0.10in}
\end{table}

\subsection{Speed and GPU Resource Comparison}
\label{sec: speed_gpu_resource_comparison}

We compare the inference speed and GPU memory consumption between our Baton and the previous joint video-audio generation models, as shown in Table \ref{table:speed_gpu_resource_comparison}. Notably, we utilize the full version of LTX-2~\citep{hacohen2026ltx} instead of the distilled version for fair performance comparison. 
Compared to our DiT backbone Ovi~\citep{low2025ovi}, Baton introduces approximately 28\% additional GPU memory from the VA-Planner's autoregressive planning and extra cross-attention layers, while achieving over 36\% improvement in DeSync and 78\% in P-Acc, demonstrating that the planning overhead is a highly favorable trade-off. Compared to MOVA~\citep{team2026mova}, Baton uses 11\% less GPU memory and is $5\times$ faster while delivering substantially better quality across all metrics, indicating that explicit semantic planning is a more efficient path to cross-modal coordination than simply scaling model and data.

\begin{table}[t!]\small
\caption{User preference of Baton compared to other competitors. A higher score indicates users prefer more to our model.
}
\vspace{-0.10in}
\begin{center}
\renewcommand\arraystretch{1.1}
\scalebox{0.85}{
\begin{tabular}{l|ccccc}
\toprule
Model    & V-A    & A-A    & I-C    & B-C    & V-A-S  \\ \midrule
Ovi~\citep{low2025ovi}      & 94.8\% & 95.3\% & 94.1\% & 94.4\% & 93.5\% \\
JavisGPT~\citep{liu2025javisgpt} & 95.2\% & 96.4\% & 95.0\% & 94.9\% & 96.6\% \\
UniAVGen~\citep{zhang2025uniavgen} & 96.7\% & 97.2\% & 98.3\% & 95.5\% & 96.2\% \\
LTX-2~\citep{hacohen2026ltx}    & 90.4\% & 88.6\% & 91.8\% & 90.2\% & 89.4\% \\
MOVA~\citep{team2026mova}     & 93.5\% & 94.2\% & 91.6\% & 89.4\% & 91.0\% \\ \bottomrule
\end{tabular}
}
\end{center}
\label{table:user_study}
\vspace{-0.10in}
\end{table}

\subsection{User Study Details}
\label{sec: user_study_details}

\begin{figure}[t!]
\begin{center}
\includegraphics[width=0.9\linewidth]{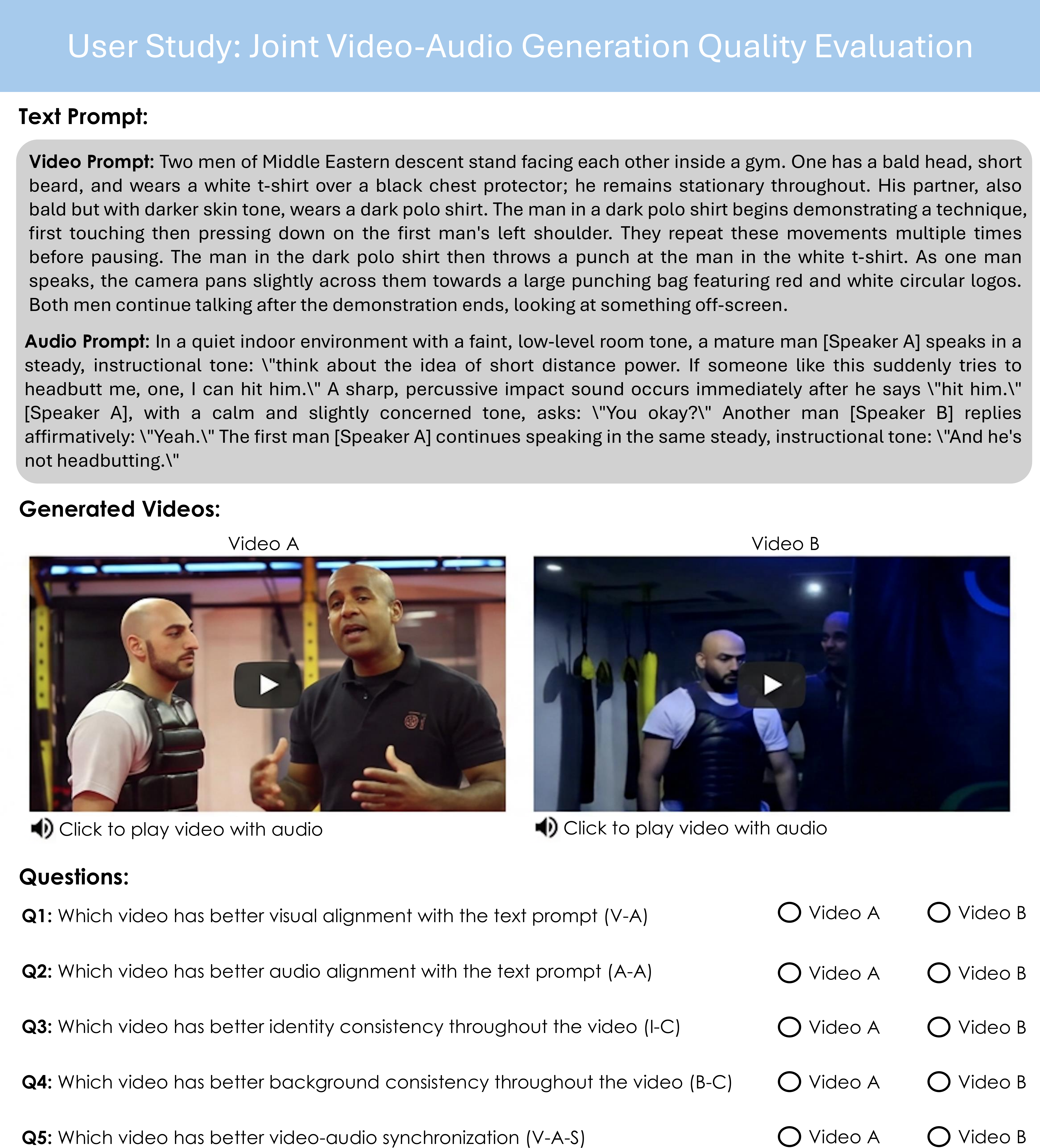}
\end{center}
\vspace{-0.2cm}
   \caption{The user study screenshot.
   }
\label{fig:user_study_screenshot}
\vspace{-0.45cm}
\end{figure}

Table \ref{table:user_study} shows the user study results. Fig. \ref{fig:user_study_screenshot} illustrates a screenshot of our user study.
The 30 test videos were randomly sampled from our Sem100 dataset, which consists of 100 unseen videos (10 seconds long) collected from diverse social media platforms (YouTube, BiliBili). 
These videos span various ethnicities, genders, and indoor/outdoor settings.

A total of 200 individuals took part in this evaluation. Eligibility was restricted to adults aged 18 and older who possessed normal or corrected-to-normal visual acuity and reported no auditory deficits. Participants were recruited through internal university mailing lists and campus announcements at our institution. The participant pool drew predominantly from the university's students (approximately 75\%), with the remainder comprising faculty members (25\%). Disciplinary backgrounds varied widely, including computer science, engineering, and the arts. The sample maintains an approximately even gender balance.
The evaluation followed a two-alternative forced choice (2AFC) paradigm. On every trial, text prompts are first displayed to the participant, after which two videos appear in a randomized sequence: one produced by Baton and the other by a competitor. Participants indicated which of the two videos they judged superior along five criteria: video alignment with text prompts, audio alignment with text prompts, identity consistency, background consistency, and video-audio synchronization. Every participant assessed the full set of 30 cases, with the left–right ordering of the two clips randomized independently per trial to mitigate positional bias.
All responses are gathered through an online survey interface. Preference rates are then computed for every method pairing by aggregating judgments across the entire participant pool and all cases.
In terms of IRB approval, we confirm that we consulted our university's institutional ethics board before conducting the study.
The study was granted an exemption from full IRB review under the institution's minimal-risk research provisions, based on three conditions.
First, the study involved only the perceptual comparison of pre-generated video outputs and did not involve any intervention, deception, or interaction beyond viewing videos and selecting preferences. 
Second, no personally identifiable information (such as names, email addresses, or demographic details beyond aggregate statistics) was collected or stored. Third, participation was entirely voluntary, and participants were informed of the study's purpose and their right to withdraw at any time before beginning. 
Each participant generally takes 10 minutes to accomplish our user study, and has received 5 dollars for compensation.

\subsection{Multi-Speakers Results}
\label{sec: multiple_speaker_results}
\begin{figure}[t!]
\begin{center}
\includegraphics[width=1\linewidth]{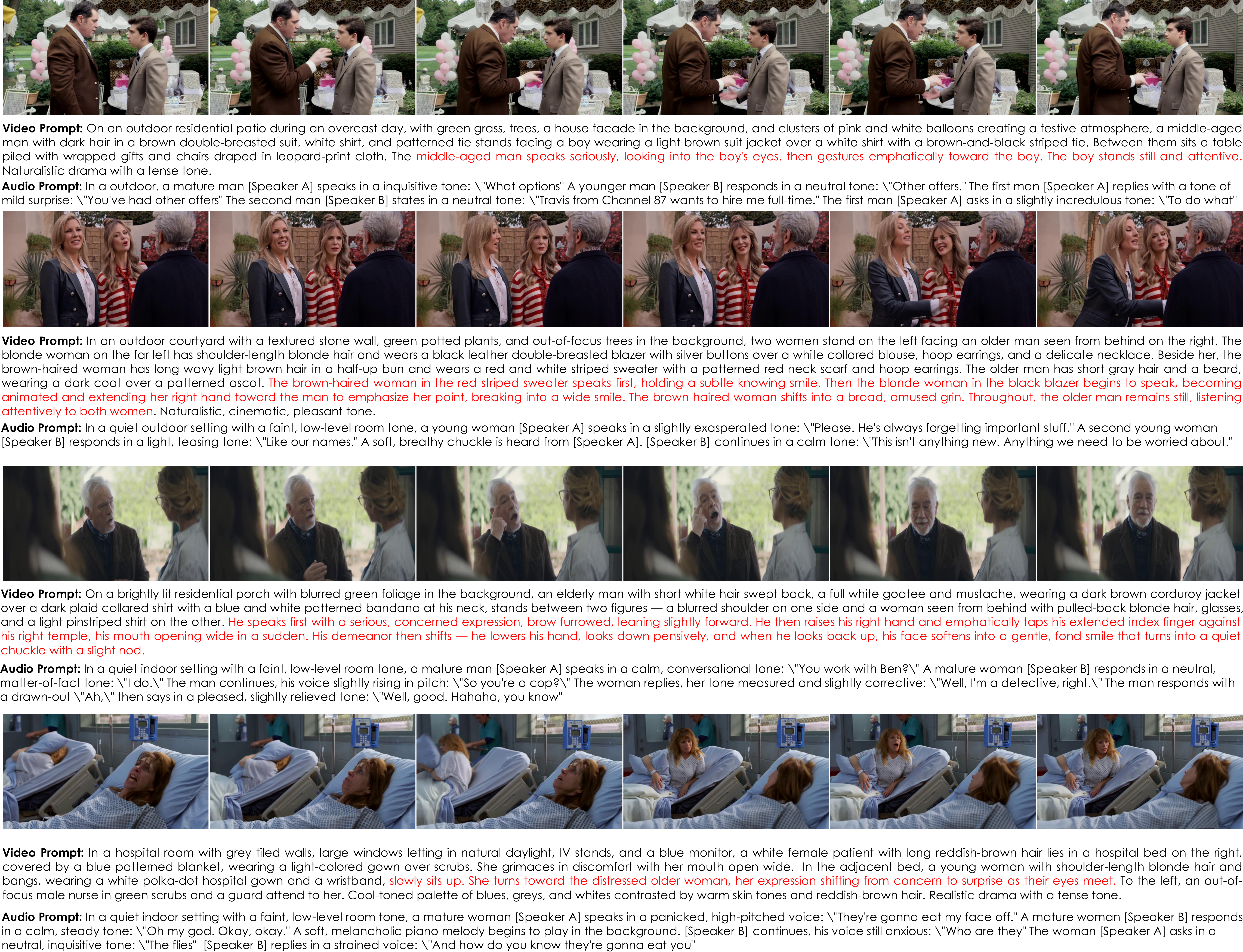}
\end{center}
\vspace{-0.3cm}
   \caption{Synthesized video-audio content involving multi-speakers. Please refer to the demo video for audio.
   }
\label{fig:multi_speaker}
\vspace{-0.25cm}
\end{figure}

Fig.\ref{fig:multi_speaker} shows some multi-speaker cases. The results demonstrate that Baton is capable of handling complex scenarios involving multi-person interaction or communication.

\subsection{Cartoon Video Results}
\label{sec: cartoon_video_results}
\begin{figure}[t!]
\begin{center}
\includegraphics[width=1\linewidth]{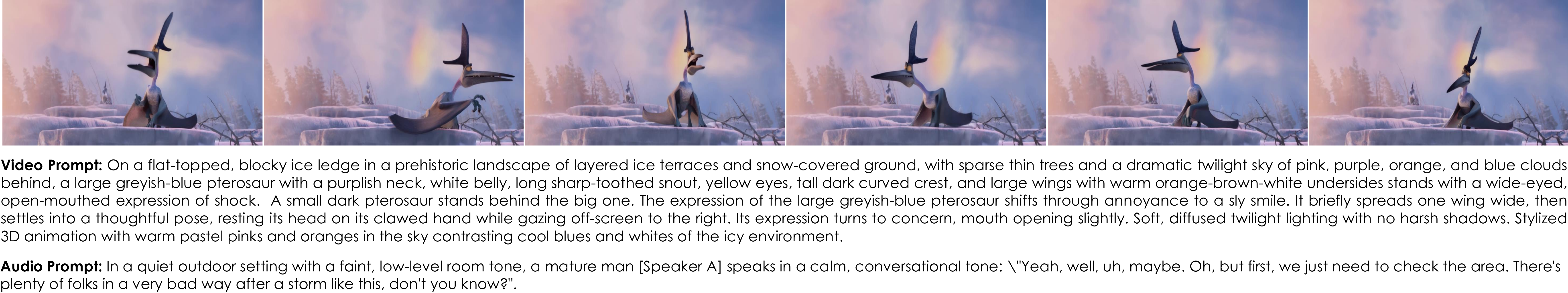}
\end{center}
\vspace{-0.3cm}
   \caption{Cartoon video-audio content. Please refer to the demo video for audio.
   }
\label{fig:cartoon}
\vspace{-0.25cm}
\end{figure}

To validate the diversity of Baton, we experiment on a cartoon case, as shown in Fig. \ref{fig:cartoon}. The results indicate that Baton can also synthesize cartoon or virtual videos, highlighting its superiority in diversity.

\subsection{Complex Scene Results}
\label{sec: complex_scene_results}

\begin{figure}[t!]
\begin{center}
\includegraphics[width=0.95\linewidth]{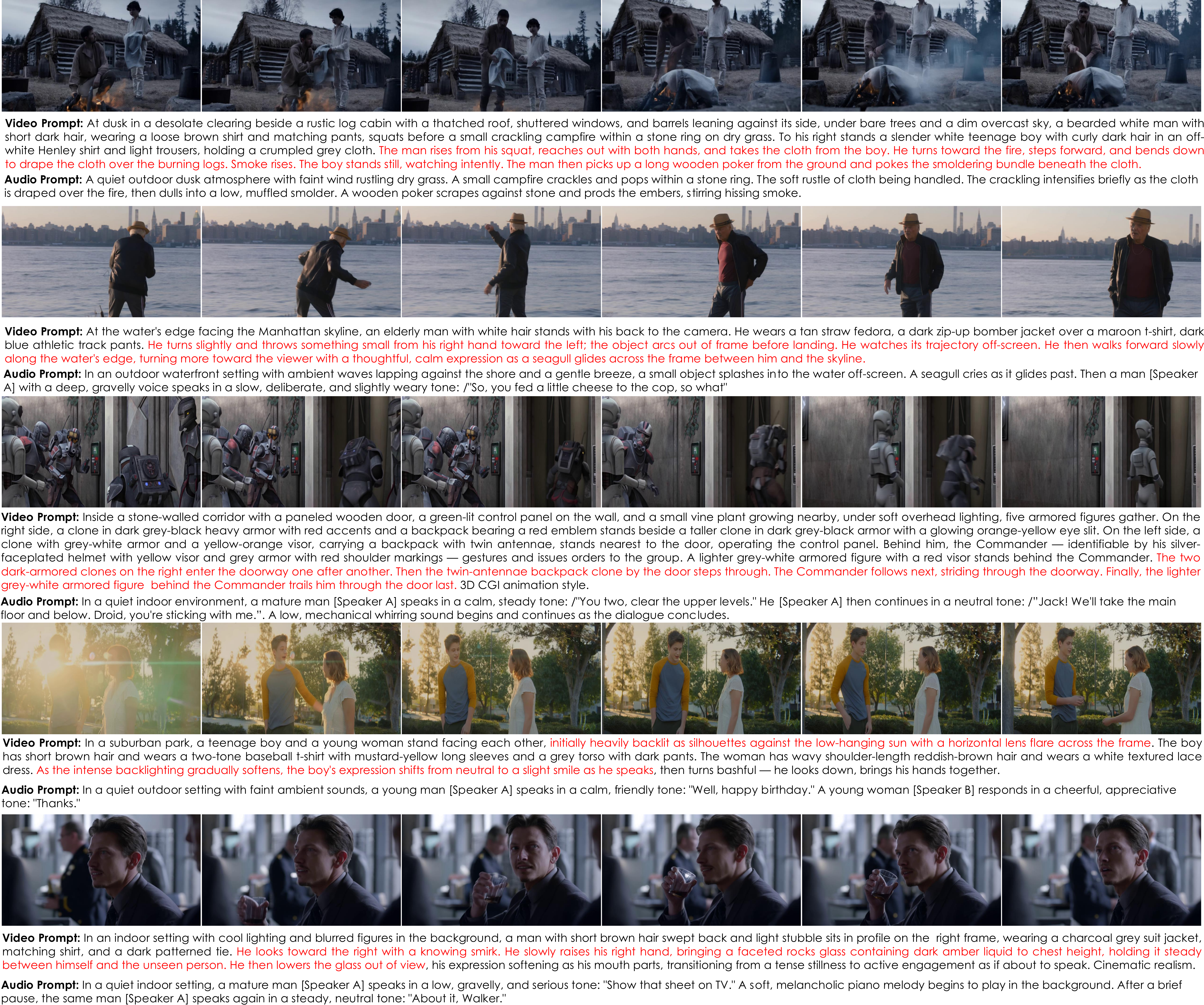}
\end{center}
\vspace{-0.3cm}
   \caption{Complex scene results (1/5). Please refer to the demo video for audio.
   }
\label{fig:complex_scene_1}
\vspace{-0.25cm}
\end{figure}

\begin{figure}[t!]
\begin{center}
\includegraphics[width=0.95\linewidth]{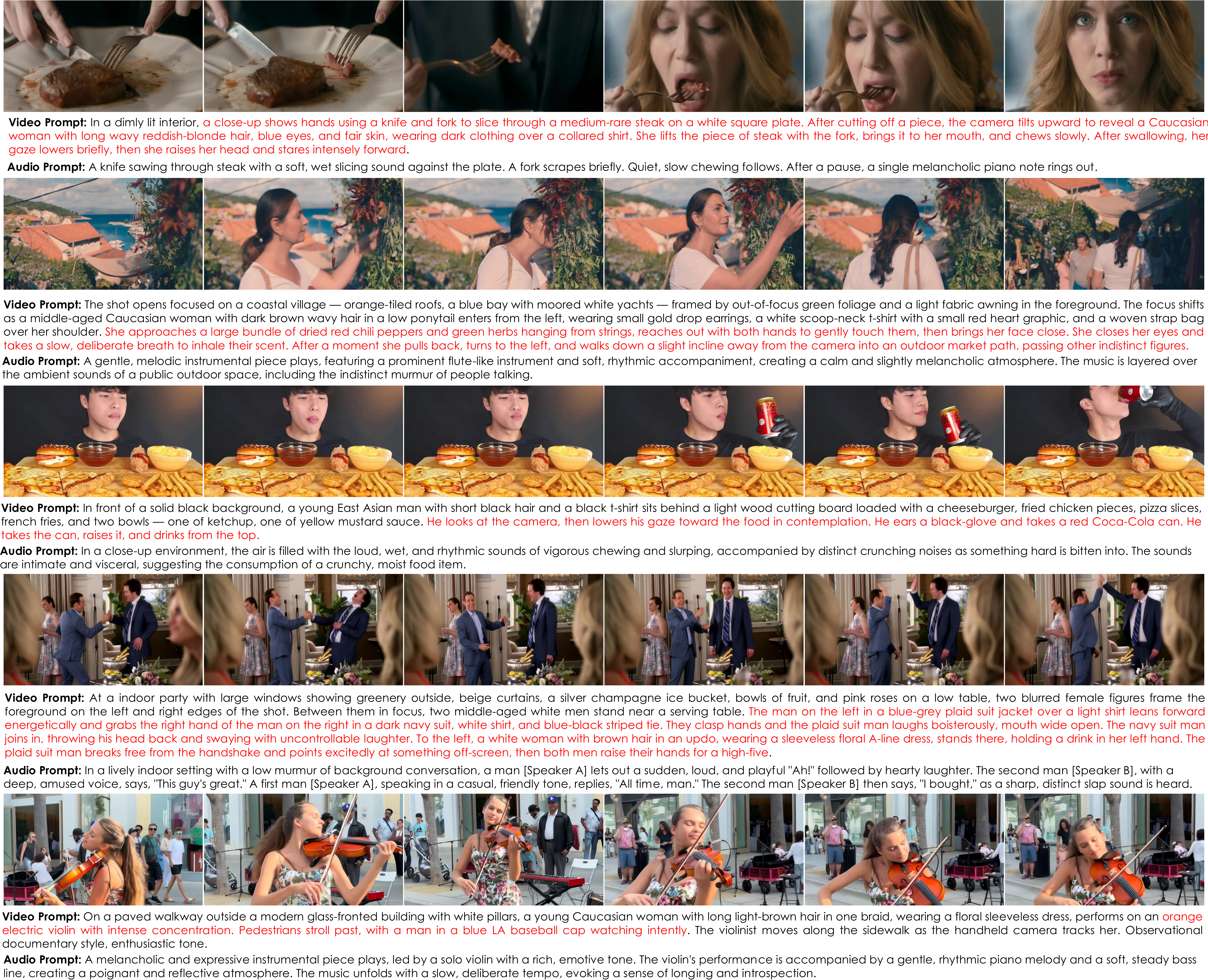}
\end{center}
\vspace{-0.3cm}
   \caption{Complex scene results (2/5). Please refer to the demo video for audio.
   }
\label{fig:complex_scene_2}
\vspace{-0.25cm}
\end{figure}

\begin{figure}[t!]
\begin{center}
\includegraphics[width=0.95\linewidth]{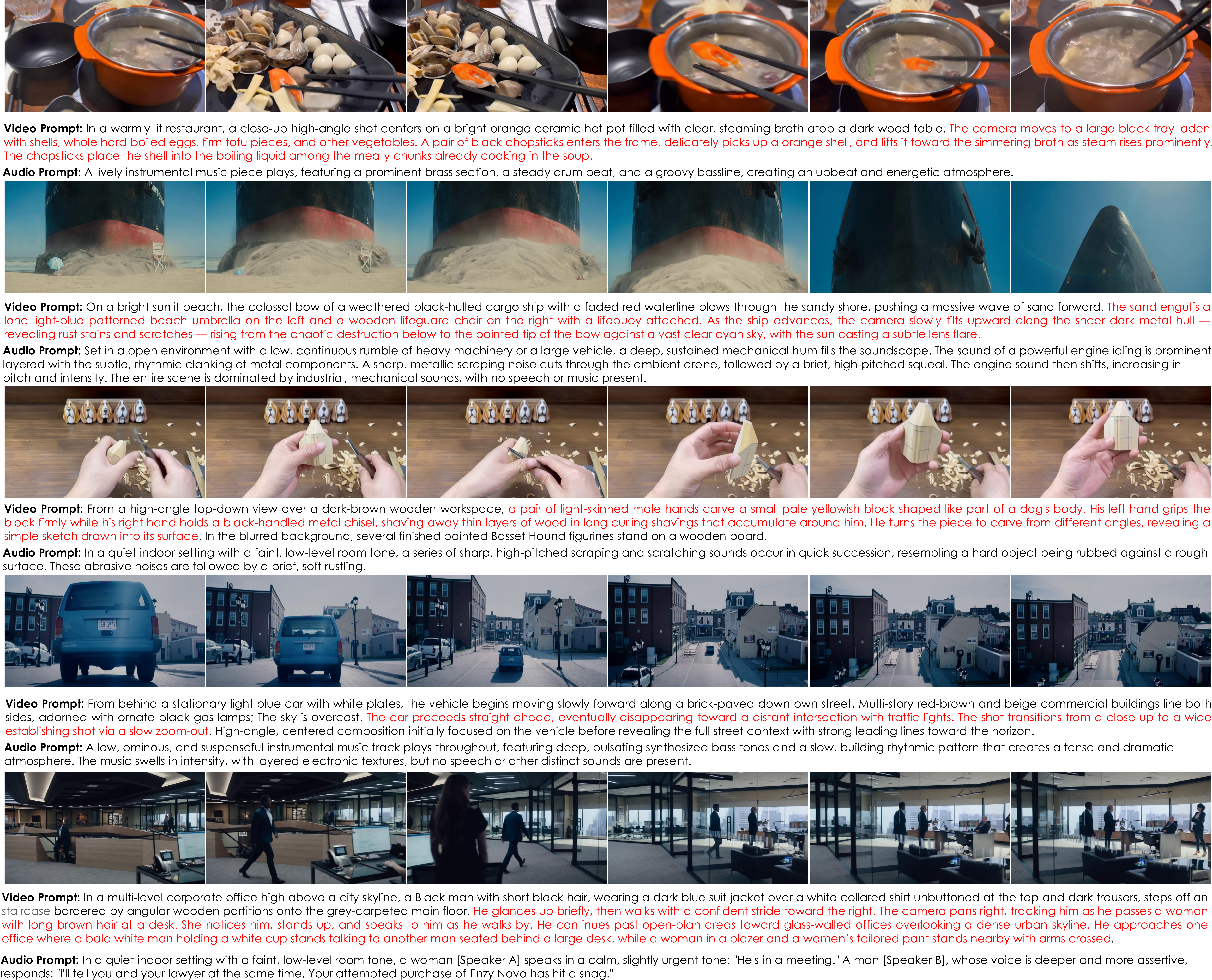}
\end{center}
\vspace{-0.3cm}
   \caption{Complex scene results (3/5). Please refer to the demo video for audio.
   }
\label{fig:complex_scene_3}
\vspace{-0.25cm}
\end{figure}

\begin{figure}[t!]
\begin{center}
\includegraphics[width=0.95\linewidth]{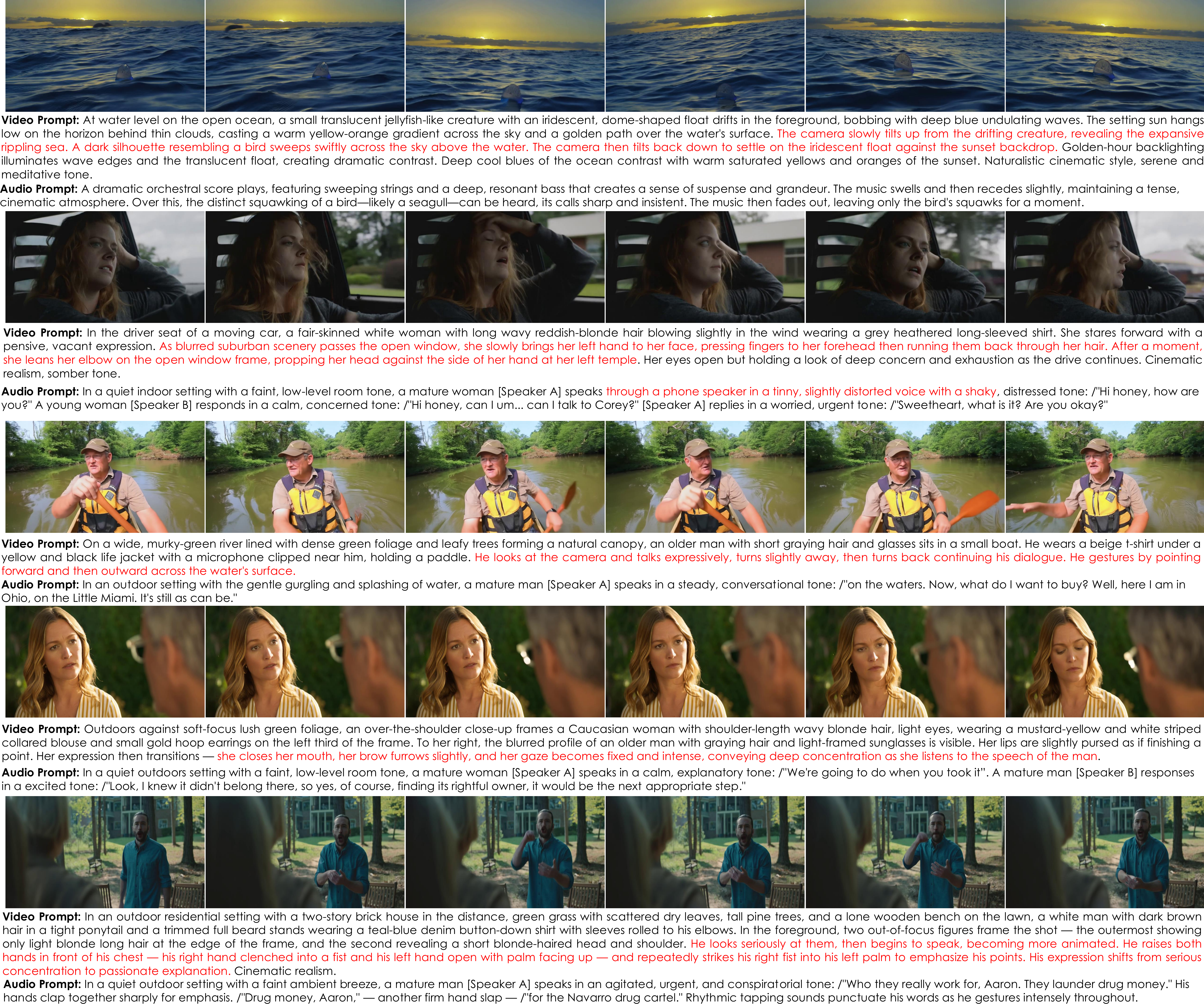}
\end{center}
\vspace{-0.3cm}
   \caption{Complex scene results (4/5). Please refer to the demo video for audio.
   }
\label{fig:complex_scene_4}
\vspace{-0.25cm}
\end{figure}

\begin{figure}[t!]
\begin{center}
\includegraphics[width=0.95\linewidth]{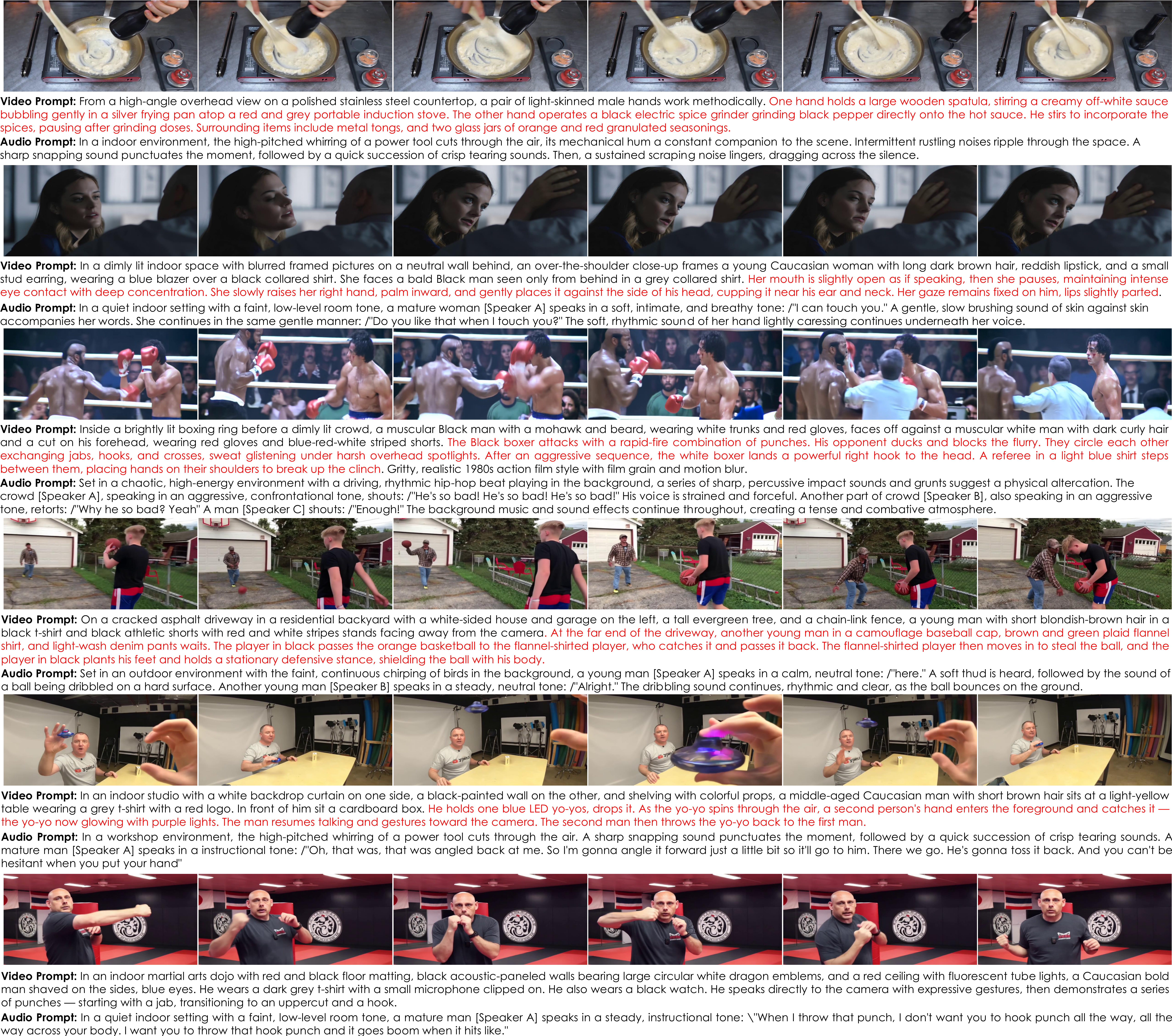}
\end{center}
\vspace{-0.3cm}
   \caption{Complex scene results (5/5). Please refer to the demo video for audio.
   }
\label{fig:complex_scene_5}
\vspace{-0.25cm}
\end{figure}

Fig. \ref{fig:complex_scene_1}, Fig. \ref{fig:complex_scene_2}, Fig. \ref{fig:complex_scene_3}, Fig. \ref{fig:complex_scene_4}, and Fig. \ref{fig:complex_scene_5} show the complex scene results.
The test cases encompass complex scenarios requiring semantic reasoning, including sequences of multiple narrative events, continuous interactions between characters and their environment or objects, logically coherent camera movements, large-scale interactions (motion) among multiple characters, evolving natural scenes, ordered multi-event actions involving specific individuals, and environment-interaction sounds that adhere to physical laws.
For example, the first, second, and fourth rows in Fig. \ref{fig:complex_scene_1} demonstrate narrative-driven character-environment interactions with physically plausible ambient sounds that respect the acoustic properties of the scene.
The third row in Fig. \ref{fig:complex_scene_1} and the fourth row in Fig. \ref{fig:complex_scene_2} showcase ordered multi-event sequences involving specific individuals, where each action must be completed before the next begins in the correct temporal order.
The fifth row in Fig. \ref{fig:complex_scene_2} and the first row and the third row in Fig. \ref{fig:complex_scene_3} present complex narrative-driven human-object interactions that require reasoning over causal dependencies between actions and their outcomes.
The first row in Fig. \ref{fig:complex_scene_2} and the fourth and fifth rows in Fig. \ref{fig:complex_scene_3} feature logically coherent camera movements that follow the narrative progression of the scene.
Fig. \ref{fig:complex_scene_5} contains cases with intense multi-character interaction dynamics, instructional narrative descriptions, and interpersonal interactions between characters.
We can observe that our Baton can perform a wide range of joint video-audio generation while accurately following the complex user prompt description and preserving the protagonist’s appearance, background, identity, and synchronization between video and audio.

\subsection{Ethics Concerns}
\label{sec: ethics_concern}

Our Baton can synthesize joint video-audio content based on the given text prompts, which can be implemented in digital human creation and film creation. However, Baton carries risks of misuse, including the creation of deepfake videos for identity impersonation, non-consensual manipulation of a person's likeness, and the spread of misinformation through fabricated speech videos on social media.
To mitigate this, it is essential to integrate visible and invisible watermarking into all generated content to ensure traceability during public deployment, incorporating automated sensitive content detection and moderation pipelines (e.g., deepfake detectors) before any output is released, restricting access through API-level authentication and usage logging, and establishing clear terms of service that prohibit generating content of real individuals without their explicit consent.

\subsection{Limitations and Future Work}
\label{sec: limitation}

\begin{figure}[t!]
\begin{center}
\includegraphics[width=1\linewidth]{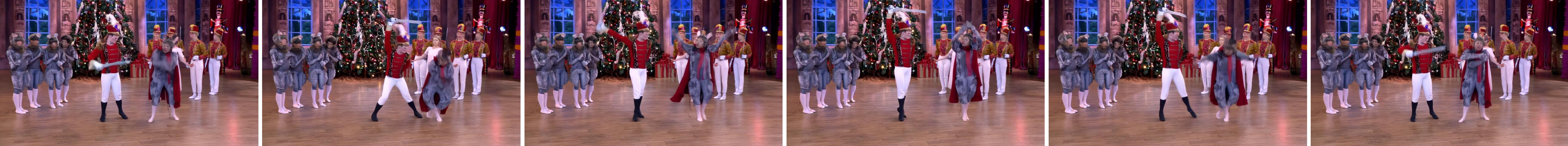}
\end{center}
\vspace{-0.2cm}
   \caption{One failure case of our Baton.
   }
\label{fig:limitation}
\vspace{-0.45cm}
\end{figure}

Fig. \ref{fig:limitation} shows one failure case of our Baton. When generating scenes involving multiple characters where each individual occupies a relatively small spatial region in the frame, the synthesized faces tend to appear blurry and lack fine-grained details. This is because the planned tokens are predicted at a sparse keyframe resolution with limited spatial tokens per frame, and when characters are small, only a few spatial tokens cover each face region, providing insufficient semantic guidance for the DiT to reconstruct detailed facial features. 
Moreover, the underlying 3D VAE compresses spatial dimensions aggressively, further reducing the effective resolution available for small faces. 
One potential solution is to incorporate a dedicated face refinement module or a super-resolution stage that operates on detected face regions, or to adopt an adaptive spatial token allocation strategy that assigns denser planned tokens to regions containing small faces.
This part is left as future work. 

\begin{figure}[t!]
\begin{center}
\includegraphics[width=0.8\linewidth]{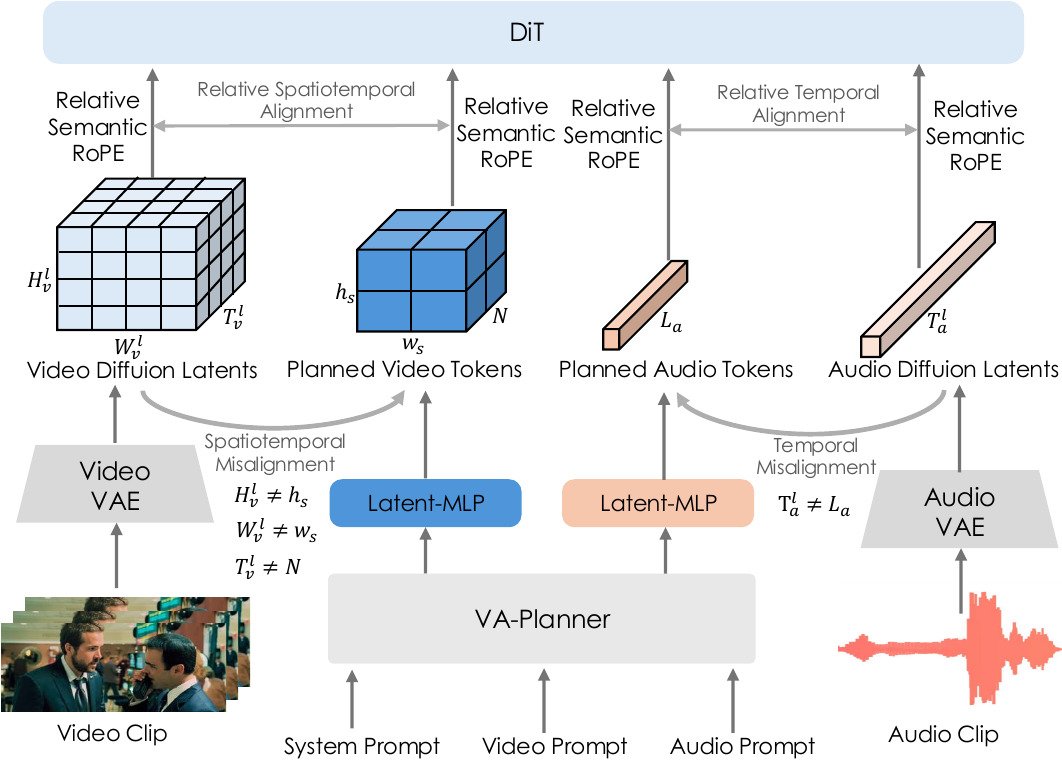}
\end{center}
\vspace{-0.3cm}
   \caption{The motivation of Relative Semantic RoPE).
   }
\label{fig:rope_motivation}
\vspace{-0.25cm}
\end{figure}

\begin{figure}[t!]
\begin{center}
\includegraphics[width=1\linewidth]{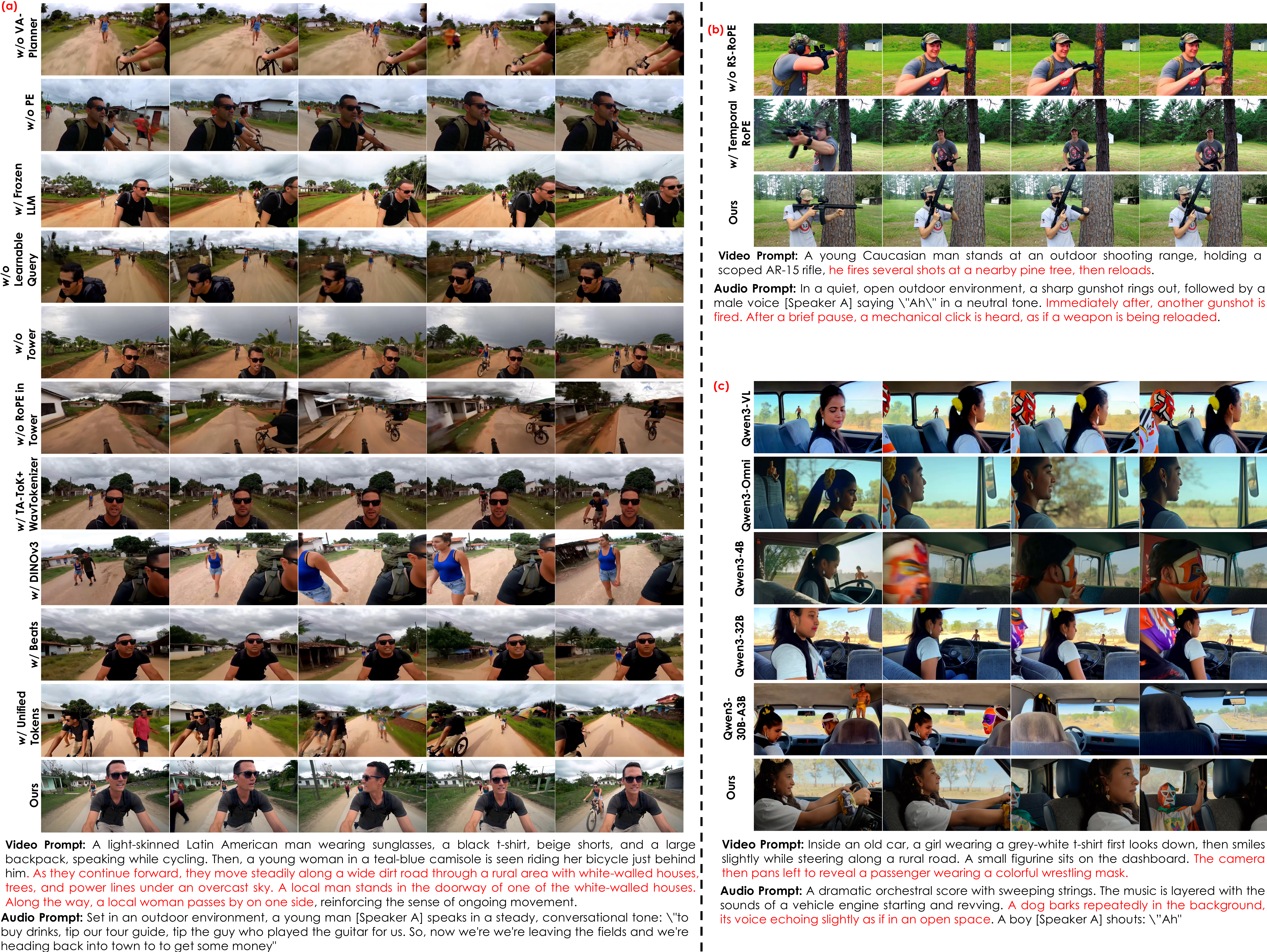}
\end{center}
\vspace{-0.3cm}
   \caption{Ablation study on VA-Planner (a), RS-RoPE (b), and different backbones (c).
   }
\label{fig:ablation}
\vspace{-0.45cm}
\end{figure}

\end{document}